\documentclass[journal]{IEEEtran}
\usepackage{cite}
\usepackage{amsmath,amssymb,amsfonts}
\usepackage{algorithmic}
\usepackage{graphicx}

\usepackage{times}
\usepackage{epsfig}
\usepackage{graphicx}
\usepackage{amsmath}
\usepackage{amssymb}
\usepackage{epstopdf}
\usepackage{booktabs}
\usepackage{subfigure} 
\usepackage{balance}
\usepackage{multirow}
\usepackage{algorithm}
\usepackage{algorithmic,eqparbox,array}

\usepackage{url}
\usepackage{graphics}

\usepackage{hyperref}
\hypersetup{
    colorlinks=true,
    linkcolor=blue,
    citecolor=green,
    filecolor=magenta,
    urlcolor=cyan,
    anchorcolor=blue,
}

\usepackage{url}            
\usepackage{graphics}

\newcommand{\etal}{{\em et al.\,}}       
\newcommand{\eg}{{\em e.g.}}           
\newcommand{\ie}{{\em i.e.}}           
\newcommand{\re}[1]{\textcolor{black}{#1}}
\newcommand{\remr}[1]{\textcolor{black}{#1}}

\usepackage{textcomp}

\renewcommand{\baselinestretch}{1}

\def\BibTeX{{\rm B\kern-.05em{\sc i\kern-.025em b}\kern-.08em
    T\kern-.1667em\lower.7ex\hbox{E}\kern-.125emX}}
\begin{document}
\title{Inconsistency-aware Uncertainty Estimation for Semi-supervised Medical Image Segmentation}
\author{Yinghuan~Shi,
        Jian~Zhang, Tong~Ling, Jiwen~Lu, Yefeng~Zheng, Qian~Yu, Lei~Qi, Yang~Gao
\thanks{Yinghuan Shi, Jian Zhang, Tong Ling and Yang Gao are with the State Key Laboratory for Novel Software Technology, Nanjing University, China. They are also with National Institute of Healthcare Data Science, Nanjing University, China. E-mail: syh@nju.edu.cn, zhangjian7369@smail.nju.edu.cn, lt@nju.edu.cn, gaoy@nju.edu.cn.}
\thanks{Jiwen~Lu is with the Department of Automation, Tsinghua University, China. E-mail: lujiwen@tsinghua.edu.cn.}
\thanks{Yefeng~Zheng is with Tencent Jarvis Lab, China. E-mail: yefengzheng@tencent.com.}
\thanks{Qian Yu is with School of Data and Computer Science, Shandong Women’s University, China. E-mail: yuqian@sdwu.edu.cn.}
\thanks{Lei Qi is with the Key Laboratory of Computer Network and Information Integration (Ministry of Education), School of Computer Science and Engineering, Southeast University, Nanjing 211189, China, and also with the State Key Laboratory for Novel Software Technology, Nanjing University, Nanjing 210046, China (e-mail: qilei@seu.edu.cn).}
}

\maketitle

\begin{abstract}
In semi-supervised medical image segmentation, \re{most previous works draw on the common assumption} that higher entropy means higher uncertainty. In this paper, we investigate a novel method of estimating uncertainty. We observe that, when assigned different misclassification costs in a certain degree, if the segmentation result of a pixel becomes inconsistent, this pixel shows a relative uncertainty in its segmentation.
Therefore, we present a new semi-supervised segmentation model, namely, conservative-radical network (\textit{CoraNet} in short) based on our uncertainty estimation and separate self-training strategy. In particular, our \textit{CoraNet} model consists of three major components: a conservative-radical module (CRM), a certain region segmentation network (C-SN), and an uncertain region segmentation network (UC-SN) that could be alternatively trained in an end-to-end manner. We have extensively evaluated our method on various segmentation tasks with publicly available benchmark datasets, including CT pancreas, MR endocardium, and MR multi-structures segmentation on the ACDC dataset. Compared with the current state of the art, our \textit{CoraNet} has demonstrated superior performance. In addition, we have also analyzed its connection with and difference from \re{conventional methods of uncertainty estimation} in semi-supervised medical image segmentation.
\end{abstract}


\begin{figure}[htbp]
\centering
\includegraphics[width=3in]{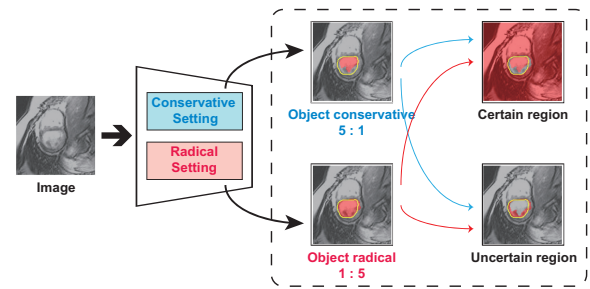}
\caption{The proposed method to estimate certain and uncertain regions. By separately training object conservative and radical models using relative misclassification cost (positive \textit{vs}. negative) of 1:5 and 5:1 respectively, the region with the consistent prediction between these two models is regarded as the certain region, otherwise it is regarded as the uncertain region. The yellow contour indicates the ground truth.}
\label{fig:definition}
\end{figure}

\section{Introduction}\label{sec:introduction}
\IEEEPARstart{P}{recise} boundary segmentation of organs and tissues of human body on medical images (\eg, computed tomography (CT), magnetic resonance imaging (MRI) and ultrasound) remains a significant yet challenging task in image processing and computer vision research field. Yet, segmentation on medical images is greatly different from segmentation on natural images, as medical images are subjected to many factors, \eg, large shape variation, low image contrast, and irregular body movement within intra- and inter-subjects, all posing considerable challenges to robust segmentation. Recently, many deep learning-based segmentation models have been proposed~\cite{unet2015MICCAI}\cite{wang2017central}\cite{gu2019net}\cite{yu2019crossbar}, as a result of which the segmentation accuracy has now been greatly improved.

Despite the renovation of previous segmentation models, most of them adopt the fully supervised setting, namely, each image in the training set is associated with manual delineation as ground truth before training the model. However, as we know, in the supervised setting, precise pixel-wise manual delineation remains a very time-consuming and laborious task. Besides, given the massive number of images, the total delineation time would be extremely long. Moreover, some inevitable subjective factors (\eg, carelessness and fatigue) would also affect the quality of manual delineation~\cite{YuCrossover2020}. In this sense, this supervised setting largely hinders the deployment of segmentation models in many clinical scenarios.

Whereas in semi-supervised learning, it is easy to assess large quantity of unlabeled images without manual delineation, so several attempts have been made to develop semi-supervised segmentation models to alleviate the requirement of large quantities of labeled (delineated) images~\cite{sun2020teacherstudent}\cite{Chen2019miccai}\cite{Ta2020MICCAI}\cite{bai2017semi}\cite{yu2019uncertainty}.
We notice that most previous semi-supervised segmentation models consist of two phases: 1) \emph{pseudo label initialization}---training an initial segmentation model only on the available labeled training samples before applying the obtained initial model to predict the pseudo label (\ie, initial segmentation) for the unlabeled samples; and 2) \emph{joint training of labeled and unlabeled samples}---jointly training a final semi-supervised segmentation model on both the labeled samples along with their ground truth, and unlabeled samples along with their pseudo label. Although these previous methods have utilized various loss functions and network structures, \re{most of them} follow \re{the same method of estimating ``uncertainty" for unlabeled samples---\textbf{\emph{higher entropy value of the pixel/voxel indicates higher uncertainty}}}. \re{Note that, though there are other uncertainty estimation methods in deep learning models, \eg, Bayesian modeling, dropout and input augmentation, we notice that in semi-supervised segmentation tasks, most previous semi-supervised methods~\cite{sun2020teacherstudent}\cite{Chen2019miccai}\cite{Ta2020MICCAI}\cite{bai2017semi}\cite{yu2019uncertainty} tend to adopt the simple uncertainty estimation method via setting a predefined threshold on the output of softmax layer, as it is easy to implement. For simplicity, we refer to this conventional way of estimation as \textit{\remr{the softmax-based confidence}} in the following parts.}

Despite the promising performance \re{of conventional methods of semi-supervised medical image segmentation} on different tasks, we observe that the following two issues should be addressed.

\textbf{Observation 1:} ``\emph{\re{It is the first step that matters the most}}''. Under the semi-supervised setting, the initial pseudo label prediction for unlabeled samples plays a crucial role in the process of segmentation. The more errors it introduces at the beginning of training, the more error propagations it might cause in the proceeding process. Therefore, how to ensure a safe and reliable model initialization becomes critical. Most previous methods estimate ``uncertainty" with \remr{softmax-based confidence} by seeking a trade-off between precision and recall. \textit{Can we investigate a novel method of estimating ``uncertainty"}? 

\textbf{Observation 2:} ``\emph{Not all the regions will be equally treated}''. In previous semi-supervised segmentation, for unlabeled samples, usually both the certain region (with high confidence) and the uncertain region (with low confidence) are input to the same segmentation network, which might not only render it hard to make full use of certain region, but also underestimate the complexity of the uncertain region. \textit{Can we separately treat their segmentation in a unified framework}?

To this end, in this paper we investigate the estimation of \textit{uncertainty} from a new perspective. Considering the fact that when assigned different misclassification costs on classes, the predicted label of a pixel during segmentation becomes inconsistent (to a certain degree), so its current prediction seems not to be certain enough (see Figure \ref{fig:definition}). Hence, to 1) provide a robust estimation of uncertainty and 2) separately segment certain and uncertain regions, we present a novel framework namely \underline{\textbf{Co}}nservative-\underline{\textbf{Ra}}dical network (\textit{\textbf{CoraNet}}) for semi-supervised segmentation.

Our CoraNet model consists of three major components: 1) a conservative-radical module (\textbf{CRM}), which involves and maintains a main segmentation model, to indicate the certain and uncertain region masks by predicting the inconsistency between different misclassification costs; 2) a certain region segmentation network (\textbf{C-SN}) to update the model with predicted certain regions; and 3) an uncertain region segmentation network (\textbf{UC-SN}) to segment the predicted uncertain region and update the model. These three components could be alternatively trained in an end-to-end manner.
Overall, the contributions of our work include:
\begin{itemize}
\item A framework of semi-supervised segmentation with a new method of estimating pixel-wise uncertainty;
\item A novel conservative-radical module to automatically identify uncertain regions according to inconsistency prediction between different misclassification costs;
\item A training strategy to separately tackle the segmentation on certain and uncertain regions in different ways.
\end{itemize}

Our model outperforms the baselines and several state-of-the-art methods with higher accuracy on various medical image segmentation tasks, including CT pancreas, MR endocardium, and MR ACDC multi-structures segmentation. \re{We have released our implementation at \url{https://github.com/koncle/CoraNet}.}


\section{Related Work}
\label{sec:relatedwork}

\subsection{Medical Image Segmentation}
In these years, convolutional neural networks have achieved state-of-the-art (SOTA) performance in medical image segmentation. \re{In terms of the segmentation under supervised setting, among the current SOTA methods, the most popular framework is the fully convolutional network (FCN) \cite{long2015fully} based encoder-decoder architecture with the representative model like U-Net \cite{unet2015MICCAI}. So far, to make the conventional encoder-decoder architecture more effective and robust, researchers have made great efforts in the following three directions: 1) developing novel structures, including the 3D structure, recurrent neural network (RNN) based model and cascaded framework \cite{milletari2016v}\cite{CHEN2018446}\cite{Bai16Cascaded}\cite{Shi17MICCAI}\cite{christ2017automatic}\cite{yu2019crossbar}, 2) designing novel network blocks, including attention mechanism, dense connection, inception or multi-scale fusion \cite{Chen2018TMIseg}\cite{8932614}\cite{ChenISBIYan19}\cite{Wang2020AAAI}\cite{Nazir2019TIP}, and 3) utilizing sophisticated loss functions  \cite{WongMICCAI2018}\cite{ChenCVPR2019}\cite{Davood2019}\cite{YuCrossover2020}, significantly improving segmentation accuracy.}

\subsection{Uncertainty Estimation}
\re{
In deep neural network models, a reliable estimation of uncertainty plays a crucial role in quantifying the confidence of predictions. However, as the ``ground truth" uncertainty is usually not available, uncertainty estimation remains a challenging task.}

\re{So far, several methods have been developed for uncertainty estimation, especially by the way of supervised learning. For example, in Bayesian uncertainty modeling based methods, the posterior distribution over parameters is calculated on the training samples with Bayesian neural networks to estimate the uncertainty. Since it is intractable to compute exact Bayesian inference, there would be several approximate variants \cite{louizos16unc}\cite{JosUNC15}\cite{CharlesUNC15} of Bayesian neural networks for efficient training. For example, Monte Carlo dropout was proposed to perform dropout at test time to estimate the uncertainty \cite{gal16unc}. An ensemble-based estimation method was presented in \cite{BalajiUncertanty17}, while data augmentation based methods were also utilized to estimate the uncertainty by investigating the predictions under different augmentation operators \cite{Ayhan2018TesttimeDA}\cite{wang2019aleatoric}. Moreover, uncertainty estimation also triggered some interests in medical image analysis \cite{Alanunc19}\cite{Marcunc20}\cite{Zachunc18}. However, our estimation method substantially differs from previous uncertainty estimation methods that mainly adopt supervised learning with sufficient training samples. Most previous works either 1) utilize predefined thresholds for determining certain/uncertain regions or 2) measure the difference of output in different initialization or augmentations. By contrast, our goal is to estimate uncertainty by employing different costs without any prerequisite, \eg, choosing specific augmentation operators. Therefore, we believe that our way of uncertainty estimation, which has never been investigated before, is novel.
}

\subsection{Semi-supervised Learning}
Recently, with the development of deep learning, many deep learning-based semi-supervised learning (SSL) methods have been developed. For example, $\Pi$-model \cite{laine2016temporal} adopts a self-ensembling strategy by imposing a consistency prediction constraint between two augmentations of the same sample. Another self-ensembling method, \ie, temporal ensembling \cite{laine2016temporal}, extends $\Pi$-model by considering the network predictions over previous epochs.
In the mean-teacher model \cite{tarvainen2017mean}, a teacher model is maintained by averaging the model weights of consecutive student models, which was a simple yet effective method in SSL. Miyato \etal\cite{miyato2018virtual} proposed a virtual adversarial training (VAT) model by introducing the adversarial training with virtual labels in semi-supervised scenarios. Moreover, given recent progress in data augmentation 
\cite{Cubuk19autoaugmentation}\cite{Cubuk20randaugmentation}, several methods have been proposed to apply data augmentation in semi-supervised learning, \eg, MixMatch \cite{Ber19mixmatch} and FixMatch \cite{sohn2020fixmatch}.
In summary, most of recent state-of-the-art SSL models adopted 1) feature-based or image-based augmentation to benefit the learning from unlabeled data, and 2) specifically designed methods (\eg, temporal ensembling) to reduce the possible noise in unlabeled data.

\begin{table}[htbp]
\footnotesize
\centering
\caption{\re{The comparison of recent methods. For simplicity, we use EM and CR to indicate the categories of entropy minimization and consistent regularization, respectively.}}
\label{tab:related_works}
\renewcommand\arraystretch{0.95}
\begin{tabular}{l|ccc}
\toprule
\multirow{2}*{Method} & Treat region & \multicolumn{2}{c}{Modeling unlabeled data}  \\
& separately? & \scriptsize{\textit{High conf. reg.}} & \scriptsize{\textit{Low conf. reg.}} \\
\midrule
\cite{kervadec2019curriculum} \scriptsize{(MICCAI'19)} & - & EM & EM \\
\cite{yu2019uncertainty} \scriptsize{(MICCAI'19)} & - & CR & CR \\
\cite{li2020transformation} \scriptsize{(TNNLS'21)} & - & CR & CR \\
\cite{KulluriICCV19} \scriptsize{(ICCV'19)} & - & EM & EM \\
\cite{Zhou2020miccai} \scriptsize{(MICCAI'20)} & - & CR & CR \\
\cite{Fang2020DMNetDM} \scriptsize{(MICCAI'20)} & - & CR & CR \\
\cite{Ouali_2020_CVPR} \scriptsize{(CVPR'20)} & - & CR & CR \\
\cite{Chen2019miccai} \scriptsize{(MICCAI'19)} & - & CR & CR \\
\cite{luo2020semisupervised} \scriptsize{(AAAI'21)} & - & CR & CR \\
\cite{peng2020mutual} \scriptsize{(MIDL'20)} & - & CR & CR \\
\cite{XIA2020101766} \scriptsize{(MedIA'20)} & - & CR & CR \\
\midrule
\textbf{Ours} & $\surd$ & EM & CR \\ 
\bottomrule
\end{tabular}
\end{table}

\subsection{Semi-supervised Medical Image Segmentation}
To alleviate the challenge of manual delineation, several methods have been recently proposed to improve segmentation performance in the semi-supervised environment. \re{These previous methods usually fall into either entropy minimization (EM) based methods or consistency regularization (CR) based methods.}
For example, Bai \etal \cite{bai2017semi} proposed a method namely semiFCN that performs self-training for cardiac MR image segmentation by integrating labeled and unlabeled data during training. 
Kervadec \etal \cite{kervadec2019curriculum} presented a segmentation framework based on inequality constraints, which could tolerate uncertainties during segmentation in inferred knowledge in semi-supervised manner.
Yu \etal \cite{yu2019uncertainty} developed a novel uncertainty-aware semi-supervised framework for left atrium segmentation on MR images, 
while Li \etal \cite{li2020transformation} employed the mean teacher-based framework to optimize the network by a weighted combination of common supervised loss and regularization loss.
Sun \etal \cite{sun2020teacherstudent} developed a hierarchical attention module with a teacher model to output the pseudo labels and a student model to combine manual labels and pseudo labels for joint training. 
In \cite{Zhou2020miccai}, a teacher-student model was investigated by introducing perturbation-sensitive sample mining to jointly perform semantic segmentation and feature distillation.
\re{The consistency between different decoders was imposed during segmentation in \cite{Fang2020DMNetDM}\cite{Ouali_2020_CVPR}.}
\re{Huo \etal \cite{Huo_2021_CVPR} developed an iterative training scheme in semi-supervised segmentation manner.}
Luo \etal \cite{luo2020semisupervised} presented a dual-task-consistency semi-supervised framework with a dual-task deep network that jointly predicts segmentation maps and geometry-aware level set representation. 
\re{The mutual information based regularization and transformation consistency were integrated in \cite{peng2020mutual} for semi-supervised segmentation.} 
\re{A unified framework with uncertainty-aware modeling was developed to address both the unsupervised domain adaptation and semi-supervised segmentation tasks in \cite{XIA2020101766}.}

\re{To better summarize and compare these works, we roughly group them into EM-based or CR-based methods in Table \ref{tab:related_works}.} As is shown, despite their success, most of these works tend to estimate uncertainty by evaluating the pixel/voxel-wise entropy. Yet, since uncertainty could greatly influence the subsequent learning process, we aim to investigate the possibility of developing other safe and reliable methods of estimating uncertainty. \re{So we propose a novel way of modeling region-level uncertainty with CRM, which could help the model to separately treat different regions, a method proven effective in our evaluation.}

\textbf{Remark.}
In summary, our proposed method is substantially different from previous semi-supervised segmentation methods.
\re{Drawing on the semi-supervised learning, these previous methods usually are either EM-based or CR-based methods, so based on the EM or CR architecture, they usually introduce additional consistency-based constraints, \eg, output-based consistency \cite{Fang2020DMNetDM}\cite{Ouali_2020_CVPR}\cite{peng2020mutual}\cite{WangMICCAI20Double} or task-based consistency \cite{luo2020semisupervised}\cite{XIA2020101766}. By contrast, our method is highly novel in 1) \textit{providing a region-level mask to indicate certain/uncertain regions with our proposed CRM}, and in 2) \textit{training the subsequent segmentation module by incorporating the merits from both CR and EM based architectures}. According to our best knowledge, the aforementioned two aspects have never been investigated in previous works.}


\begin{figure*}[htbp]
\centering
\includegraphics[width=7.2in]{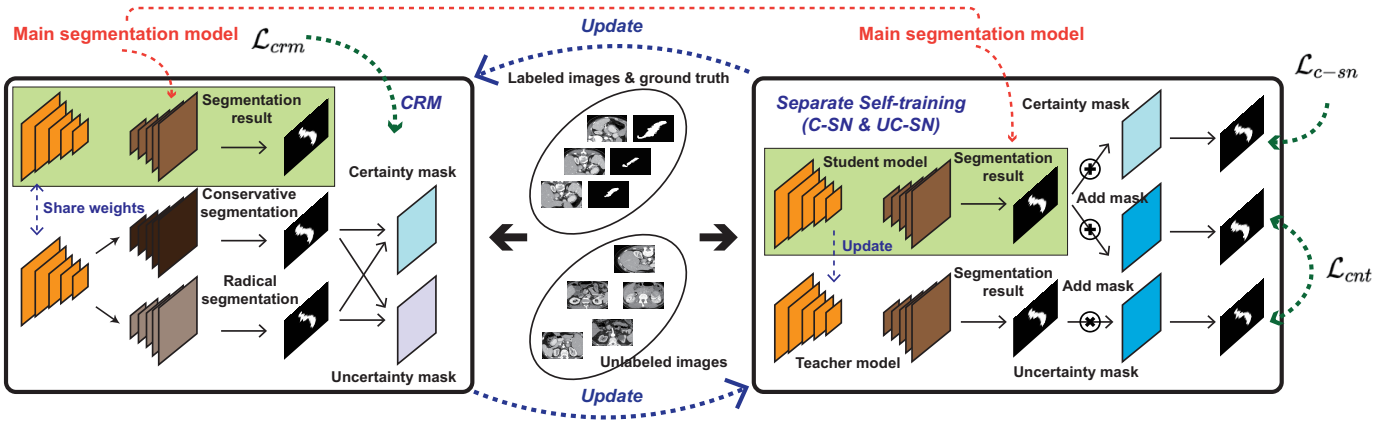}
\caption{The main framework of our proposed method \textit{CoraNet}.}
\label{fig:framework}
\end{figure*}

\section{The Proposed Method}
\color{black}
\label{sec:ourmethod}
 
\subsection{Problem Analysis}

\color{black}

For clear clarification, we show a visualized example in Figure \ref{fig:definition} to illustrate the case of segmentation with just two classes---each pixel is classified into \emph{background} (\ie, \emph{class 0}) or \emph{object} (\ie, \emph{class 1}) for simplicity. For the loss function, we employ the popularly used cross-entropy loss for illustration. Before providing our definition of uncertainty, we introduce two cost-sensitive settings as follows:
\begin{itemize}
\item \textbf{Object conservative setting} (\eg, \textbf{5:1}): The cost of \emph{mis-segmenting one background pixel to object class} equals to that of \emph{mis-segmenting five object pixels to background class}. Obviously, it performs segmentation in an object-caution manner, since in most case only the most reliable central part of the object is segmented. In this setting, the predicted object class under object conservative setting is highly confident.
\item \textbf{Object radical setting} (\eg, \textbf{1:5}): We reverse the misclassification cost in object conservative setting. The cost of \emph{mis-segmenting one object pixel to background class} equals to that of \emph{mis-segmenting five background pixels to object class}. It is a background-caution setting that indicates the predicted background is highly confident.
\end{itemize}

On the same training set, by employing these two cost-sensitive settings to train two separate segmentation models, as shown in Figure \ref{fig:definition}, we can obtain the corresponding object-conservation and object-radical models.\footnote{We will use \emph{conservative} and \emph{radical} for simplicity in the following.} With the obtained models, we then input the same image to these two models to obtain corresponding conservative and radical segmentation masks, respectively. We calculate the difference between these two segmentation masks to define:
\begin{itemize}
\item \textbf{Certain region}: the pixels with the same prediction under both conservative and radical settings, since they are being robust to different costs.
\item \textbf{Uncertain region}: the pixels with different predictions under conservative and radical settings, indicating their current predictions are uncertain.
\end{itemize}

We now extensively analyze the ways in which our estimation method largely differs from traditional \remr{softmax-based confidence} estimation methods. From Figure \ref{fig:definition}, we can observe that most part of the image falls into the certain region, however uncertainty is also possible in non-boundary location.
Our definition of uncertainty requires neither task-specific assumption nor empirical predefined condition, \eg, boundary-aware loss, specific convolution or pooling. Although it shows potential to locate the uncertain region, we still aim to
\begin{enumerate}
\item \textit{leverage the high-confidence certain region to continuously complement the current labeled samples}, and 
\item \textit{reduce incorrect segmentation caused by low-confidence uncertain region}.
\end{enumerate}
Accordingly, we employ a separate self-training strategy to treat the subsequent segmentation of certain and uncertain regions in the unlabeled data separately. For the certain region, being aware of its high confident prediction, we employ its prediction as pseudo label for self-training. For the uncertain region, we aim to obtain a more robust estimation to prevent the possible error propagation.

Our overall framework is summarized in Figure \ref{fig:framework} with three components: 1) a conservative-radical module (CRM), 2) a certain region segmentation network (C-SN), and 3) an uncertain region segmentation network (UC-SN). \re{Note that, in Figure \ref{fig:framework}, though we split them for better visualization, the main segmentation model and student model are actually the same model.}

\textbf{Notation.} For a semi-supervised segmentation task, given $m$ labeled samples with manual delineation as ground-truth mask (label) and $n$ unlabeled samples with no ground-truth information, we denote the labeled samples as $\mathbf{X}_1^l$, $\mathbf{X}_2^l$, ..., $\mathbf{X}_m^l\in \mathbb{R}^{H_0 \times W_0}$ and their corresponding label maps as $\mathbf{Y}_1^l$, $\mathbf{Y}_2^l$, ..., $\mathbf{Y}_m^l\in \mathbb{B}^{H_0 \times W_0}$. The unlabeled samples are denoted as $\mathbf{X}_1^u$, $\mathbf{X}_2^u$, ..., $\mathbf{X}_n^u\in \mathbb{R}^{H_0 \times W_0}$. Note that, $H_0$ and $W_0$ indicate the height and width of original images, respectively. It is worthwhile to mention that, in the following description and evaluation, we first train a model on both labeled and unlabeled samples and then deploy the trained model to segment the unseen samples without any intersection to previous observed labeled/unlabeled samples, so it falls into an inductive rather than transductive semi-supervised segmentation setting.

\subsection{Certainty-aware Prediction: CRM}
Recall that the conventional cross-entropy loss $\mathcal{L}_{ce}$ defined in the segmentation task is evaluated between the prediction $p_{z,k}$ of the $z$-th pixel on the $k$-th class and its corresponding label $q_{z,k}$, so we denote two separate losses as $\mathcal{L}_{ce\text{-}con}$ and $\mathcal{L}_{ce\text{-}rad}$ to indicate the aforementioned conservative and radical settings, respectively.
Specifically, for the $i$-th labeled image ($1\leq i \leq m$), by employing the conventional cross-entropy loss $\mathcal{L}_{\text{ce}}$ with different misclassification costs, we formulate $\mathcal{L}_{ce\text{-}con}$ and $\mathcal{L}_{ce\text{-}rad}$ for a binary segmentation task as follows:
\begin{displaymath}
\mathcal{L}_{ce\text{-}con}(\mathbf{X}_i^l, \mathbf{Y}_i^l; \mathbf{E}, \mathbf{D}_{con}) = -\sum_{k=1}^{2}\sum_{z=1}^{H_0\times W_0} w^{con}_k  q_{z,k} \text{log} p_{z,k},
\end{displaymath}
\begin{displaymath}
\mathcal{L}_{ce\text{-}rad}(\mathbf{X}_i^l, \mathbf{Y}_i^l; \mathbf{E}, \mathbf{D}_{rad}) =  -\sum_{k=1}^{2}\sum_{z=1}^{H_0\times W_0} w^{rad}_k  q_{z,k} \text{log} p_{z,k}.
\end{displaymath}

As we employ the encoder-decoder segmentation structure, our normal loss $\mathcal{L}_{ce}$, conservative loss $\mathcal{L}_{ce\text{-}con}$ and radical loss $\mathcal{L}_{ce\text{-}rad}$ are parameterized by a common encoder $\mathbf{E}$ with shared parameters, and three separate decoders $\mathbf{D}$, $\mathbf{D}_{con}$ and $\mathbf{D}_{rad}$, respectively. Then,  as the corresponding weights for different classes, $w^{con}_k$ and $w^{rad}_k$ are defined as follows:
$$w^{con}_k=
\begin{cases}
\alpha & k = 0\\
1 & k = 1
\end{cases} \quad \text{and} \quad
w^{rad}_k=
\begin{cases}
1 & k = 0\\
\alpha & k = 1,
\end{cases}$$
where $\alpha > 1$ indicates the high misclassification cost in current class compared with the other class. In this paper, we set $\alpha$ as 5 in the implementation.

Therefore, the total loss $\mathcal{L}_{crm}$ of CRM is calculated on all the labeled samples, which can be defined as
\begin{equation}
\begin{split}
    & \mathcal{L}_{crm}(\mathbf{E}, \mathbf{D}, \mathbf{D}_{con}, \mathbf{D}_{rad}) =  \sum_{i=1}^m \Bigg[\mathcal{L}_{ce}(\mathbf{X}_i^l, \mathbf{Y}_i^l; \mathbf{E}, \mathbf{D}) \\ + & \mathcal{L}_{ce\text{-}con}(\mathbf{X}_i^l, \mathbf{Y}_i^l; \mathbf{E}, \mathbf{D}_{con}) + \mathcal{L}_{ce\text{-}rad}(\mathbf{X}_i^l, \mathbf{Y}_i^l; \mathbf{E}, \mathbf{D}_{rad})\Bigg].
\label{eqn:crninitial}
\end{split}
\end{equation}

Next, we can obtain three separate segmentation sub-models $\mathcal{F}$, $\mathcal{F}_{con}$ and $\mathcal{F}_{rad}$ with the same encoder but separate decoders. Based on these sub-models, for the $j$-th unlabeled image, we have conservative, radical, and normal predictions as $\mathbf{Y}^{u,con}_j$, $\mathbf{Y}^{u,rad}_j$, and $\mathbf{Y}^u_j$, respectively:
\begin{equation}
    \mathbf{Y}^u_j = \mathcal{F}(\mathbf{X}^u_j; \mathbf{E},\mathbf{D})\in \mathbb{B}^{H_0 \times W_0},
\end{equation}
\begin{equation}
    \mathbf{Y}^{u,con}_j = \mathcal{F}_{con}(\mathbf{X}^u_j; \mathbf{E},\mathbf{D}_{con}) \in \mathbb{B}^{H_0 \times W_0},
\end{equation}
\begin{equation}
    \mathbf{Y}^{u,rad}_j = \mathcal{F}_{rad}(\mathbf{X}^u_j; \mathbf{E},\mathbf{D}_{rad}) \in \mathbb{B}^{H_0 \times W_0}.
\end{equation}

With obtained conservative and radical predictions, the certain and uncertain masks of the $j$-th unlabeled image, $\mathbf{M}^{u,c}_j$ and $\mathbf{M}^{u,uc}_j$ can be defined as
\begin{equation}
    \mathbf{M}^{u,uc}_j = \mathbf{Y}^{u,rad}_j\oplus\mathbf{Y}^{u,con}_j \in \mathbb{B}^{H_0 \times W_0},
    \label{eqn:ucmaskcomp}
\end{equation}
\begin{equation}
    \mathbf{M}^{u,c}_j = \mathbf{1} - \mathbf{M}^{u,uc}_j \in \mathbb{B}^{H_0 \times W_0},
    \label{eqn:cmaskcomp}
\end{equation}
where $\oplus$ is pixel-wise XOR operator to indicate the inconsistency between $\mathbf{Y}^{u,con}_j$ and $\mathbf{Y}^{u,rad}_j$, and $\mathbf{1}$ is an all-one matrix.

\subsection{Separate Self-training: C-SN and UC-SN}
We treat the certain region and the uncertain region differently.
For the certain region indicated by $\mathbf{M}^{u,c}_j$, regarding its high confident prediction, we implement \textbf{C-SN} by using its prediction as the pseudo label for self-training. For the uncertain region indicated by $\mathbf{M}^{u,uc}_j$, given its prediction might be unreliable in current stage, we utilize \textbf{UC-SN} by employing an advanced model, \ie, \textit{mean teacher}, for label assignment.

\subsubsection{\textbf{C-SN for certain region segmentation}}
For the certain region indicated by $\mathbf{M}^{u,c}_j$, we directly use the prediction as their pseudo label and self-train the main segmentation model parameterized by encoder $\mathbf{E}$ and decoder $\mathbf{D}$ in the next iteration. Regarding its consistent prediction with both conservative and radical losses, we believe that it has high confidence on its prediction, which could aid the learning in the next round with two benefits: 1) it provides efficient training in that the model could directly assign the label to most parts of image, as most regions usually fall into the certain region, and 2) the certain region could be used to complement the labeled samples to aid robust training particularly when the labeled samples are limited.

Specifically, the objective loss of C-SN is formulated as:
\begin{equation}
\begin{split}
   \mathcal{L}_{c\text{-}sn}(\mathbf{E},\mathbf{D}) = \frac{1}{n}\sum_{j=1}^n\mathcal{L}_{ce\text{-}m}(\mathbf{X}^{u}_j, \mathbf{Y}^{u}_j, \mathbf{M}^{u,c}_{j}; \mathbf{E},\mathbf{D}),
\end{split}
\label{eqn:csnobjv}
\end{equation}
where $\mathcal{L}_{ce\text{-}m}$ is defined as:
\begin{displaymath}
\mathcal{L}_{ce\text{-}m}(\mathbf{X}^u_j, \mathbf{Y}^u_j, \mathbf{M}^{u,c}_{j}; \mathbf{E}, \mathbf{D}) = -\sum_{k=1}^{2}\sum_{z=1}^{H_0\times W_0} \mathbf{M}^{u,c}_{j,z} q_{z,k} \text{log} p_{z,k},
\end{displaymath}
where $\mathbf{M}^{u,c}_{j,z}$ denotes the $z$-th pixel of $\mathbf{M}^{u,c}_{j}$. 


\subsubsection{\textbf{UC-SN for uncertain region segmentation}}
Due to its low confident prediction, the uncertain region could not be directly used as pseudo label, so we incorporate the deep semi-supervised learning framework proposed earlier to achieve reliable pseudo label assignment instead. In particular, we introduce the \textit{mean teacher} model into our framework, where the main segmentation model (parameterized by encoder $\mathbf{E}$ and decoder $\mathbf{D}$) is regarded as the student model. In addaition to the student model, another teacher model parameterized by encoder $\mathbf{E}'$ and decoder $\mathbf{D}'$ with the same encoder-decoder structure is also employed in the training process. 

Specifically, the student and teacher models are imposed to have a consistent prediction on unlabeled samples. The consistent loss can be formulated as follows:
\begin{equation}
\begin{split}
    \mathcal{L}_{cnt} = \frac{1}{n}\sum_{j=1}^n \Big\|\mathbf{M}^{u,uc}_j \otimes (\mathcal{F}(\mathbf{X}^{u}_j; \mathbf{E}, \mathbf{D}) - \mathcal{F}'(\mathbf{X}^{u}_j; \mathbf{E}', \mathbf{D}'))\Big\|^2,
\label{eqn:ucsnobjv}
\end{split}
\end{equation}
where $\mathcal{F}$ and $\mathcal{F}'$ denote the student segmentation model and teacher segmentation model, respectively, and $\otimes$ means element-wise multiplication.
Also, given $\mathbf{E}_t$, $\mathbf{D}_t$, $\mathbf{E}'_t$ and $\mathbf{D}'_t$ at the $t$-th iteration, the parameter update of $\mathbf{E}_{t+1}'$ and $\mathbf{D}_{t+1}'$ at the ($t$+1)-th iteration is similar to that in the conventional mean teacher model as
\begin{equation}
    \mathbf{E}'_{t+1} = \beta \mathbf{E}'_t + (1-\beta)\mathbf{E}_t, \quad
    \mathbf{D}'_{t+1} = \beta \mathbf{D}'_t + (1-\beta)\mathbf{D}_t,
    \label{eqn:teacher}
\end{equation}
where $\beta$ is the smoothing coefficient parameter.

\begin{algorithm}[t]
\caption{Proposed \textit{CoraNet}}
\label{alg:algorithm}
\hspace*{0.02in}{\textbf{\textit{Training input}}}: labeled data $\mathbf{X}^l_1$, $\mathbf{X}^l_2$, ..., $\mathbf{X}^l_m$ and their label $\mathbf{Y}^l_1$, $\mathbf{Y}^l_2$, ..., $\mathbf{Y}^l_m$, unlabeled data $\mathbf{X}^u_1$, $\mathbf{X}^u_2$, ..., $\mathbf{X}^u_n$\\
\hspace*{0.02in}{\textbf{\textit{Training output}}}: segmentation model $\mathcal{F}$ parameterized by encoder $\mathbf{E}$ and decoder $\mathbf{D}$
\begin{algorithmic}[1]
\STATE \textcolor{blue}{\texttt{$//$model initialization}} \\
\STATE $\mathbf{E}$, $\mathbf{D}$, $\mathbf{D}_{con}$, $\mathbf{D}_{rad} \leftarrow$ initialized by Eqn. (\ref{eqn:crninitial}) \\
\STATE $\mathbf{E}'$, $\mathbf{D}'$ $\leftarrow$ initialized with $\mathbf{E}$, $\mathbf{D}$ \\
\STATE $\forall j$, $\textbf{M}^{u,uc}_j$ and $\textbf{M}^{u,c}_j \leftarrow$ initialized by Eqns. (\ref{eqn:ucmaskcomp}) and (\ref{eqn:cmaskcomp}) \\
\WHILE{not converged}
\STATE \textcolor{blue}{\texttt{$//$certainty-based segmentation}} \\
\STATE Update $\mathbf{E}$ and $\mathbf{D}$ using Eqn. (\ref{eqn:csnobjv}) \\
\STATE \textcolor{blue}{\texttt{$//$uncertainty-based segmentation}} \\
\STATE Update $\mathbf{E}$ and $\mathbf{D}$ using Eqn. (\ref{eqn:ucsnobjv}) \\
\STATE \textcolor{blue}{\texttt{$//$update CoraNet model}} \\
\STATE Update $\mathbf{E}$, $\mathbf{D}$, $\mathbf{D}_{con}$, $\mathbf{D}_{rad}$ using Eqn. (\ref{eqn:crninitial}) \\
\STATE \textcolor{blue}{\texttt{$//$update mask}} \\
\STATE Update $\textbf{M}^{u,uc}_j$ and $\textbf{M}^{u,c}_j$ using Eqns. (\ref{eqn:ucmaskcomp}) and (\ref{eqn:cmaskcomp}) \\
\STATE \textcolor{blue}{\texttt{$//$update teacher model}} \\
\STATE Update  $\mathbf{E}'$ and $\mathbf{D}'$ using Eqn. (\ref{eqn:teacher})
\ENDWHILE
\end{algorithmic}
~\\
\hspace*{0.02in}\textbf{\textit{Test input}}: test image $\mathbf{X}^t_1$, ..., $\mathbf{X}^t_k$\\
\hspace*{0.02in}\textbf{\textit{Test output}}: results $\hat{\mathbf{Y}^t_1}$, ..., $\hat{\mathbf{Y}^t_k}$
\begin{algorithmic}[1]
\STATE Calculate $\forall i$, $\hat{\mathbf{Y}^t_i} \leftarrow \mathcal{F}(\mathbf{X}^t_i; \mathbf{E}, \mathbf{D})$
\end{algorithmic}
\label{alg:algsummary}
\end{algorithm}

Algorithm \ref{alg:algorithm} summarizes the overall procedure of the method we propose.

\subsection{Discussion}
\subsubsection{Network Architecture}
\re{Although our method also consists of one encoder and several decoders, it is fundamentally different from previous works \cite{Fang2020DMNetDM}\cite{Ouali_2020_CVPR} for the following reasons:}

\re{\textbf{\textit{Goals are different}}: the two decoders (\ie, $\mathbf{D}_{con}$ and $\mathbf{D}_{rad}$) in our method aim to estimate their difference in later separate self-training. In this sense, as a novel way to estimate the uncertainty, the utilization of our two decoders (\ie, $\mathbf{D}_{con}$ and $\mathbf{D}_{rad}$) could be seen as an intermediate step to generate the region-level uncertain mask before being discarded later in the test phase. 
Previous works \cite{Fang2020DMNetDM}\cite{Ouali_2020_CVPR} attempt to seek a consistent prediction between different decoders, often following the conventional way of consistency regularization in semi-supervised learning.}

\re{\textbf{\textit{Implementations are different}}: our two decoders are optimized with different objective functions (\ie, radical and conservative settings in Eqn. (\ref{eqn:crninitial})) to generate diverse results for uncertainty estimation, while previous works \cite{Fang2020DMNetDM}\cite{Ouali_2020_CVPR} usually utilize the same objective function on different decoders by either introducing different perturbations or employing data augmentations to aid consistency training.}

\subsubsection{Efficiency}
\re{Compared with the simple yet popularly-used baseline U-Net, our method is more efficient in network parameters:}

\re{\textbf{\textit{Training phase}}: Our method contains merely two additional decoders ($\mathbf{D}_{con}$ and $\mathbf{D}_{rad}$), each consisting of just three convolutional layers (\ie, two 3$\times$3 and one 1$\times$1 convolutions). In this sense, our model could avoid introducing too many additional parameters. Also, although our separate self-training involves a teacher network, its weights are indeed the moving average of the weights of the student network, therefore not introducing any additional parameters here. In addition, in separate self-training module, the student model is actually the same model as the main segmentation model in Figure \ref{fig:framework}.}

\re{\textbf{\textit{Test phase}}: It is worth noting that although our method consists of three decoders, \ie, $\mathbf{D}$, $\mathbf{D}_{con}$ and $\mathbf{D}_{rad}$, during the test phase, two of them (\ie, $\mathbf{D}_{con}$, $\mathbf{D}_{rad}$) are to be discarded later in our final model, as indicated in Algorithm 1. Therefore, it is as efficient as U-Net.}

\re{Based on the above analysis, in our test phase, the inference parameters only involve those from the main segmentation model parameterized by $\mathbf{E}$ and $\mathbf{D}$. For instance, in CT pancreas segmentation, 1) in the training phase, the parameter size of U-Net and our model is 29.6M and 29.8M, respectively, 2) in the test phase, they are both 29.6M after discarding $\mathbf{D}_{con}$ and $\mathbf{D}_{rad}$. Since our method would not introduce too many additional parameters, it is fast to perform training and testing. In particular, when implementing our model on an NVIDIA GeForce RTX 2080TI GPU, the training time of the popularly used mean teacher model \cite{tarvainen2017mean} and our model is 2.685h and 2.863h, respectively, and the test time for segmenting 1,414 slices is 18.75s and 18.85s, respectively. In this sense, although our method involves two additional decoders, they have less negative effect on efficiency in both training and test phases.}

\subsubsection{Weight Setting}
\re{For how to balance the weight of $\mathcal{L}_{c\text{-}sn}$ and $\mathcal{L}_{cnt}$, we find that setting the weighting parameter as 1:1 (\ie, setting $\mathcal{L}_{c\text{-}sn}$ and $\mathcal{L}_{cnt}$ with equal weight) could achieve satisfactory performance. This is probably because the loss normally depends on the production of 1) the number of pixels in this region and 2) the ratio of possible misclassified pixels in this region.
\begin{itemize}
    \item \textit{For certain region, the number of total pixels is relatively large while the ratio of misclassification could be small.}
    \item \textit{For uncertain region, the number of total pixels is relatively small while the ratio of misclassification could be large.}
\end{itemize}
We notice these two losses are often roughly the same in most cases. Moreover, setting an extremely large or small value to this weighting parameter certainly deteriorates the performance. To eliminate possible negative impacts that might be brought in by introducing extra hyper parameter, the weight to balance $\mathcal{L}_{c\text{-}sn}$ and $\mathcal{L}_{cnt}$ is set to 1:1 in our paper.}

\section{Experiments}
\label{sec:experiments}
\subsection{Setting}
For the implementation, we employ a four-level U-Net as our backbone network that outputs the feature map with 64 channels in the last level. In each level, there are two convolutional layers, both followed by batch normalization, ReLU and a max pooling layer. Next, we utilize the transposed convolutional layer for up-sampling. In the last level, our model contains the proposed conservative-radical module, which consists of three branches, each branch adopting two \re{3$\times$3 and one 1$\times$1 convolutional layers with batch normalization and ReLU. For self-training, we first pretrain our model with labeled data for 30 epochs and then train our model with both labeled and unlabeled data for the following 100 epochs. We relabel the unlabeled data every 5 epochs to obtain their pseudo labels. For all the 2D setting-based experiments conducted in our work, the batch size is set to 4. The smoothing coefficient parameter $\beta$ is set to 0.99.} For optimization, we employ Adam to optimize our network with the learning rate as 0.001 and betas as (0.5, 0.999) in all experiments. We implement our method using PyTorch on an NVIDIA GeForce RTX 2080TI GPU.

\subsection{Datasets}
We evaluate our method on three different medical image segmentation tasks, namely, CT pancreas segmentation, MR endocardium segmentation, and ACDC segmentation.  Now we provide the details of these three datasets.

\textbf{CT Pancreas:} This dataset \cite{Roth2016Data}\cite{roth2015deeporgan}\cite{clark2013the} is by far the largest, most popular and authoritative pancreas dataset in CT images \cite{Man19TMI}. In this dataset, all images are manually delineated by an experienced physician. In our experiment, we use a collection of 82 contrast-enhanced abdominal CT volumes in this dataset. The size of CT scan ranges from $512 \times 512 \times 181$ to $512 \times 512 \times 466$ voxels. For image pre-processing, the image intensity in the same CT image is first resized and then min-max normalized to the scale of [-100, 240] to remove some background, which is irrelevant to our segmentation goal. After that, each slice is cropped to the size of $192 \times 240$ pixels in a similar way of \cite{li2019pro}. Next, we adopt data augmentation by utilizing several simple and effective operators, \ie, 1) randomly rotating the image within -30 to 30 degrees, 2) randomly rescaling the image from 0.5x to 2x, and 3) vertical or horizontal flip.
    
\textbf{MR Endocardium:} This is a public benchmark dataset of cardiac MR sequences \cite{And2008efficient}. The MR endocardium dataset consists of 7,980 MR images collected from 33 different individuals. The image size is $256 \times 256$ pixels with the inplane pixel size of 0.9-1.6 mm and the inter-slice thickness of 6-13 mm. In this dataset, our goal is to accurately segment the boundary of the LV cavity (\ie, endocardium).
    
\textbf{ACDC Dataset:} Containing 100 cine MR sequences in total, this dataset is a public benchmark dataset of the 2017 Automated Cardiac Diagnosis Challenge (ACDC).\footnote{https://www.creatis.insa-lyon.fr/Challenge/acdc/} In every image, there are three categories to segment, \ie, right ventricle (RV), left ventricle (LV) cavities, and the myocardium (epicardial contour more specifically). We introduce this dataset to validate the effectiveness of our method for multi-class segmentation tasks. Since the image size in the original dataset is not consistent, we resize all the images into 256 $\times$ 256 pixels.

\subsection{Empirical Analysis}
\label{sec:analysis}
We investigate several important issues in our model to better understand its property. Firstly, we verify that CRM could not only obtain a reliable initial prediction, but also strike a good balance between precision and recall. Then, we analyze the separate self-training with visualized results to empirically illustrate the changing trend of uncertainty estimation results.  

\subsubsection{CRM: A Reliable Initial Prediction}
In semi-supervised segmentation, an initial segmentation model is first trained on the available labeled samples, then used to assign the pseudo label for unlabeled samples. We supspect that a reliable initial prediction on unlabeled samples is crucial for the following segmentation. A good initial prediction could help reduce the error propagation and ensure the relative stability of the learning process.

To validate this hypothesis, we compare our proposed method with traditional \remr{softmax-based confidence} initialization~\cite{sun2020teacherstudent}\cite{Chen2019miccai}\cite{Ta2020MICCAI}\cite{bai2017semi}\cite{yu2019uncertainty} in segmentation task. Specifically, we introduce three \remr{softmax-based confidence} settings with the threshold of 0.5, 0.7 and 0.9, respectively. In particular, the setting of 0.7 indicates that the pixel/voxel in the unlabeled samples, whose certainty is larger than the threshold of 0.7 (evaluated by softmax output), will be assigned by its predicted label as being foreground or background. Otherwise, we do not assign any labels to them. This is a very common setting used in current semi-supervised segmentation models. Note that the setting of 0.5 and 0.9 is performed in a similar way.

\begin{figure}[htbp]
\centering
\includegraphics[width=2.6in]{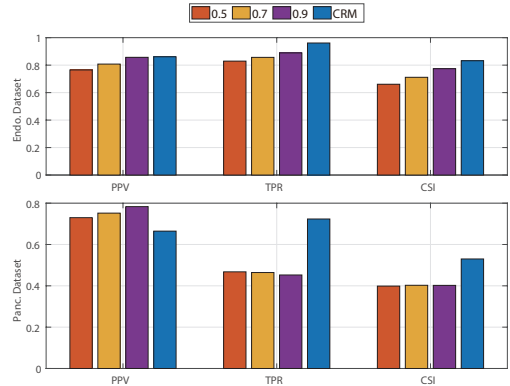}
\caption{The PPV, TPR, CSI of initial prediction on the \textit{endocardium} dataset (endo. in short) as shown on the top row and the \textit{pancreas} dataset (panc. in short) as shown on the bottom row.}\label{fig:obsv1}
\end{figure}

\begin{figure*}[htbp]
\centering
\includegraphics[width=6in]{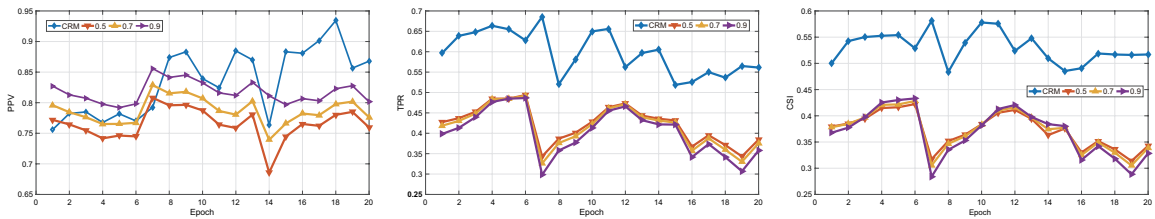}
\caption{The change of PPV, TPR, and CSI during the training on the \textit{pancreas} dataset. This validates that a good initial prediction could significantly reduce error propagation in following segmentation.}\label{fig:obsv2}
\end{figure*}

To evaluate different settings, we introduce positive predictive value (PPV, a.k.a. precision), true positive rate (TPR, a.k.a. recall), and critical success index (CSI) as evaluation metrics. Suppose \texttt{tp} indicates the true positive prediction where both ground truth and prediction are foreground; \texttt{fp} indicates the false positive prediction where the prediction is foreground and ground truth is background;
\texttt{fn} indicates the false negative prediction where the prediction is background and ground truth is foreground, the PPV, TPR and CSI are defined as
\begin{displaymath}
\text{PPV} = \frac{\texttt{tp}}{\texttt{tp}+\texttt{fp}}, \;
\text{TPR} = \frac{\texttt{tp}}{\texttt{tp}+\texttt{fn}}, \;
\text{CSI} = \frac{\texttt{tp}}{\texttt{tp}+\texttt{fp}+\texttt{fn}}.
\end{displaymath}

\begin{figure*}[htbp]
\centering
\subfigure{
\begin{minipage}[t]{0.08\linewidth}
\centering
\includegraphics[width=0.6in]{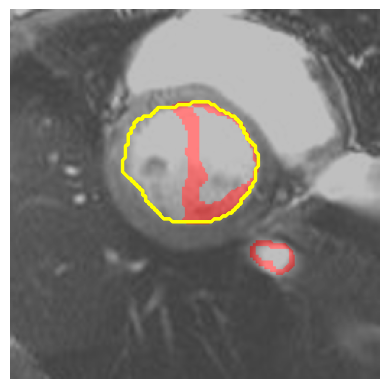}
\end{minipage}}
\subfigure{
\begin{minipage}[t]{0.08\linewidth}
\centering
\includegraphics[width=0.6in]{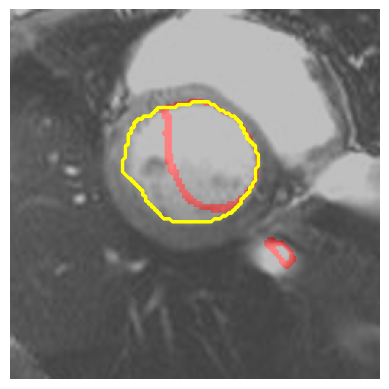}
\end{minipage}}
\subfigure{
\begin{minipage}[t]{0.08\linewidth}
\centering
\includegraphics[width=0.6in]{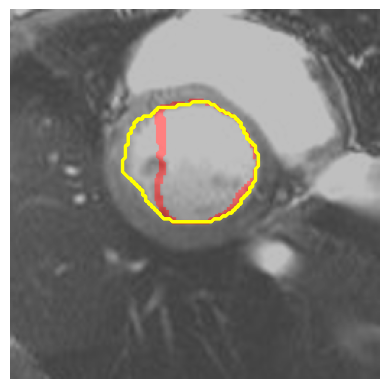}
\end{minipage}}
\subfigure{
\begin{minipage}[t]{0.08\linewidth}
\centering
\includegraphics[width=0.6in]{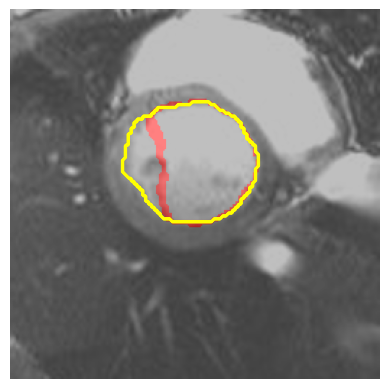}
\end{minipage}}
\subfigure{
\begin{minipage}[t]{0.08\linewidth}
\centering
\includegraphics[width=0.6in]{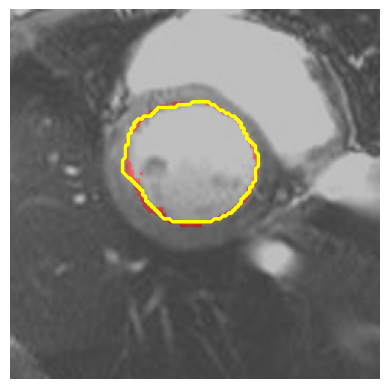}
\end{minipage}}
\hspace*{0.01in}
\subfigure{
\begin{minipage}[t]{0.08\linewidth}
\centering
\includegraphics[width=0.6in]{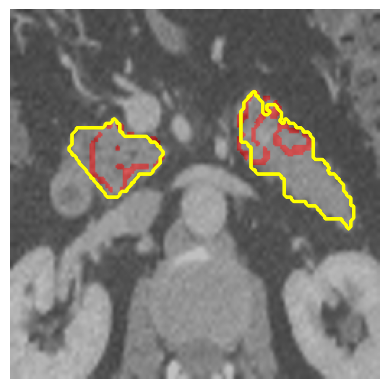}
\end{minipage}}
\subfigure{
\begin{minipage}[t]{0.08\linewidth}
\centering
\includegraphics[width=0.6in]{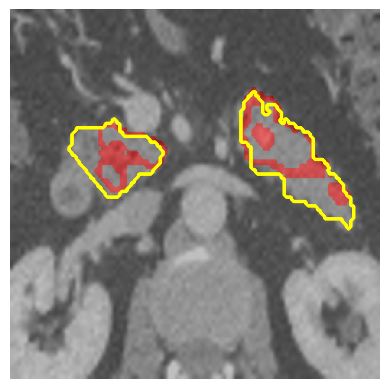}
\end{minipage}}
\subfigure{
\begin{minipage}[t]{0.08\linewidth}
\centering
\includegraphics[width=0.6in]{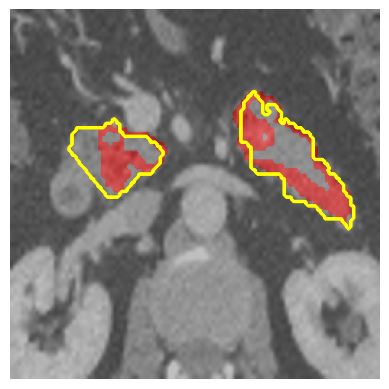}
\end{minipage}}
\subfigure{
\begin{minipage}[t]{0.08\linewidth}
\centering
\includegraphics[width=0.6in]{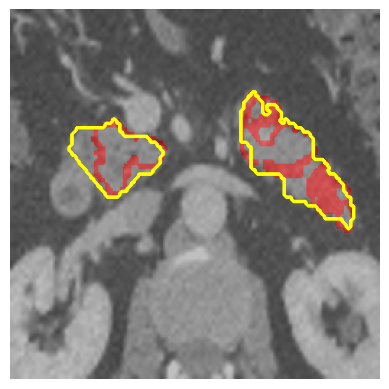}
\end{minipage}}
\subfigure{
\begin{minipage}[t]{0.08\linewidth}
\centering
\includegraphics[width=0.6in]{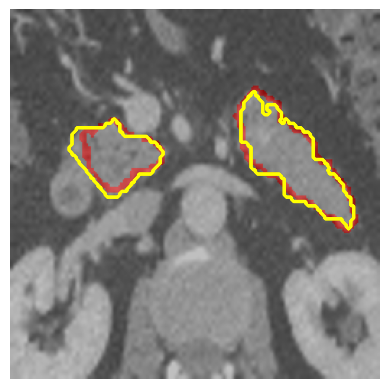}
\end{minipage}}
\\
\vspace*{-0.3cm}
\subfigure{
\begin{minipage}[t]{0.08\linewidth}
\centering
\includegraphics[width=0.6in]{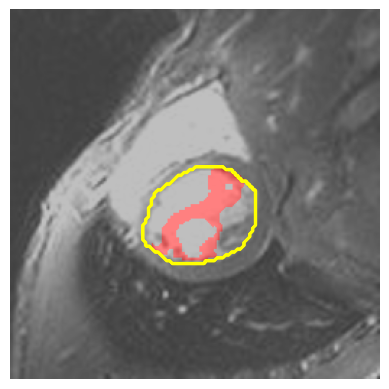}
\centerline{\scriptsize{1$^{\text{st}}$-epoch}}
\end{minipage}}
\subfigure{
\begin{minipage}[t]{0.08\linewidth}
\centering
\includegraphics[width=0.6in]{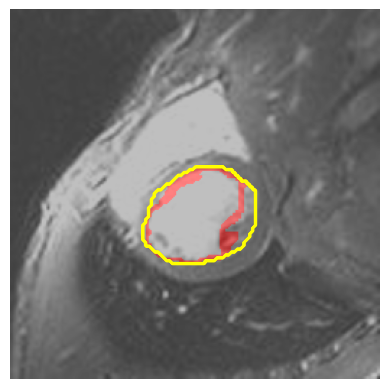}
\centerline{\scriptsize{5$^{\text{th}}$-epoch}}
\end{minipage}}
\subfigure{
\begin{minipage}[t]{0.08\linewidth}
\centering
\includegraphics[width=0.6in]{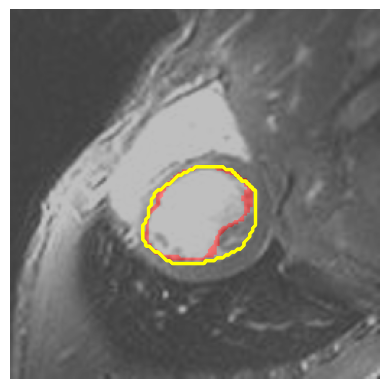}
\centerline{\scriptsize{10$^{\text{th}}$-epoch}}
\end{minipage}}
\subfigure{
\begin{minipage}[t]{0.08\linewidth}
\centering
\includegraphics[width=0.6in]{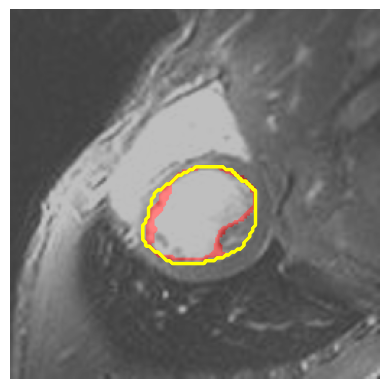}
\centerline{\scriptsize{15$^{\text{th}}$-epoch}}
\end{minipage}}
\subfigure{
\begin{minipage}[t]{0.08\linewidth}
\centering
\includegraphics[width=0.6in]{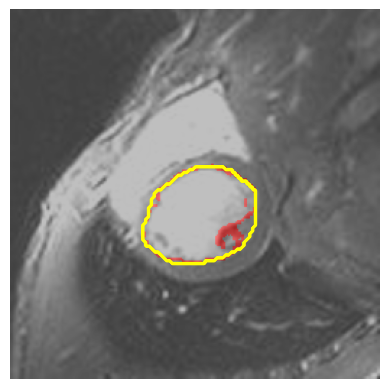}
\centerline{\scriptsize{20$^{\text{th}}$-epoch}}
\end{minipage}}
\hspace*{0.01in}
\subfigure{
\begin{minipage}[t]{0.08\linewidth}
\centering
\includegraphics[width=0.6in]{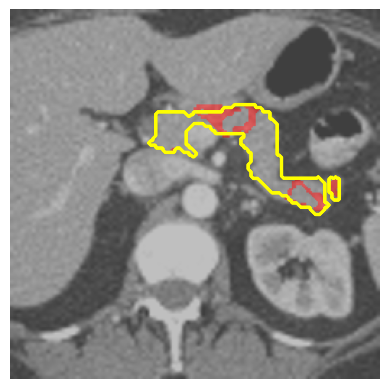}
\centerline{\scriptsize{1$^{\text{st}}$-epoch}}
\end{minipage}}
\subfigure{
\begin{minipage}[t]{0.08\linewidth}
\centering
\includegraphics[width=0.6in]{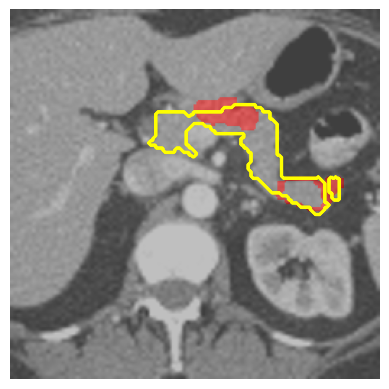}
\centerline{\scriptsize{5$^{\text{th}}$-epoch}}
\end{minipage}}
\subfigure{
\begin{minipage}[t]{0.08\linewidth}
\centering
\includegraphics[width=0.6in]{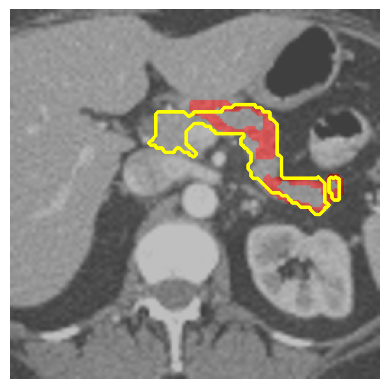}
\centerline{\scriptsize{10$^{\text{th}}$-epoch}}
\end{minipage}}
\subfigure{
\begin{minipage}[t]{0.08\linewidth}
\centering
\includegraphics[width=0.6in]{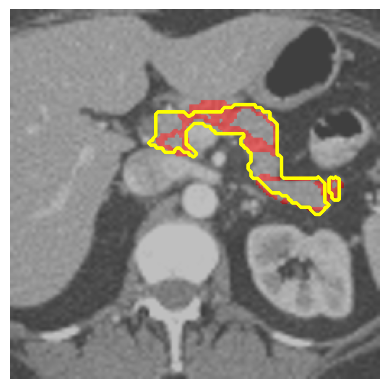}
\centerline{\scriptsize{15$^{\text{th}}$-epoch}}
\end{minipage}}
\subfigure{
\begin{minipage}[t]{0.08\linewidth}
\centering
\includegraphics[width=0.6in]{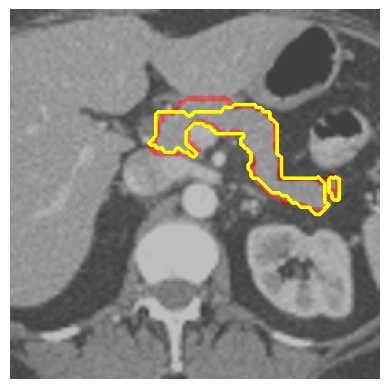}
\centerline{\scriptsize{20$^{\text{th}}$-epoch}}
\end{minipage}}
\\
\caption{Uncertain region mask predicted every five epochs on the \textit{endocardium} dataset (left) and the \textit{pancreas} dataset (right). The yellow curve indicates the ground truth, while the red region indicates the uncertainty mask.}\label{fig:exp:visualuncmask}
\end{figure*}


During segmentation, balancing the precision and recall for foreground region is crucial for accurate segmentation. Normally, higher recall indicates that more true foreground pixels are correctly classified as foreground, while higher precision means a higher proportion of true foreground pixels in classified foreground pixels.


The accuracy of initial segmentation is crucial for semi-supervised learning. To reveal how our method and the \remr{softmax-based confidence} methods balance the precision and recall in the initial segmentation, we conduct experiments on the \textit{pancreas} and \textit{endocardium} datasets, the results of which are presented in Figure \ref{fig:obsv1}. Our CRM outperforms the \remr{softmax-based confidence} methods with different thresholds (\ie, 0.5, 0.7, and 0.9) on all metrics, except the PPV on the pancreas dataset, which means more pancreas pixels are neglected in our method. Next, we simply add the pseudo labels generated by different methods into training (\ie, self-training) for pancreas segmentation.
We report the different curve changes of all these settings in Figure \ref{fig:obsv2}, which shows that although the PPV of our method is lower than these baselines at the first epoch, the PPV improves when the training epoch increases. In addition, it is easy to see that the settings of 0.5, 0.7 and 0.9 are with the similar trend where the setting of 0.5 shows better performance in TPR and the setting of 0.9 shows better performance in PPV. 

This result validates our empirical observation that \textit{the different settings of the entropy-loss-based approach only seek a similar trade-off between precision and recall}. Compared with these entropy-loss-based approaches, our method has a better trade-off between precision and recall, indicating a large deviation to the trend of existing threshold-based setting. This experiment further discloses that \textit{a good initial prediction could significantly reduce error propagation in later segmentation} (please refer to the second and third figures in Figure \ref{fig:obsv2}). 

\subsubsection{Separate Self-training: Changing trend of Uncertainty}
In our model, we utilize a separate self-training approach to modeling the segmentation on certain and uncertain regions in different ways. We investigate if separate self-training could boost the performance compared with mean-teacher and self-training. Taking the \textit{pancreas} dataset as an example, we use half of all the images as labeled samples and the other half as unlabeled samples. According to our evaluation, the mean Dice scores of separate self-training, mean teacher and self-training are $67.01\%$, $61.23\%$ and $62.33\%$, respectively, which validate the superior performance of our separate self-training. For more details, please refer to our experimental results.

We also present several examples to visually show the changing trend of uncertainty mask every five epochs during our separate self-training in Figure \ref{fig:exp:visualuncmask}. We illustrate our observation on both the \textit{endocardium} and \textit{pancreas} datasets. In this figure, the yellow curve indicates the ground truth and the red region indicates the uncertainty mask. It could be observed that the uncertainty mask changes during the training process. At the beginning, the uncertainty mask was usually located at hard regions even in the central part of foreground, as indicated by inconsistent prediction made by CRM. As the training epoch increases, when the model shows its increasing confidence on previous uncertain region, the uncertainty mask would be located at some boundary region. This empirical observation reveals that \textit{the uncertain region might change with the evolution of the segmentation model}. Since uncertainty might change, it also reminds us that we could not always blindly treat boundary region as the uncertain region.

\subsection{Results on CT Pancreas}
We now present our evaluation on CT pancreas segmentation task.
For the baselines, we introduce popularly used \textbf{U-Net} \cite{unet2015MICCAI}, as well as semi-supervised segmentation models: \ie, \textbf{$\Pi$-model} \cite{laine2016temporal}, \textbf{temporal ensembling} \cite{laine2016temporal} (\textbf{TE} in short), \textbf{mean teacher} \cite{tarvainen2017mean} (\textbf{MT} in short) and uncertainty-based mean teacher (\textbf{UA-MT} in short) \cite{yu2019uncertainty}. \re{For the implementation of these baseline methods, we refer to a publicly available implementation of U-Net.\footnote{\re{https://lmb.informatik.uni-freiburg.de/people/ronneber/u-net/.}} We also utilize the implementation of {$\Pi$-model},\footnote{\re{https://github.com/smlaine2/tempens.}} TE\footnote{\re{https://github.com/smlaine2/tempens.}} and MT\footnote{\re{https://github.com/CuriousAI/mean-teacher.}} available in their literature and replace the networks with U-Net for our segmentation task. For UA-MT, we directly adopt a publicly available implementation.\footnote{\re{https://github.com/yulequan/UA-MT.}}}
Specifically, since U-Net is a fully supervised segmentation model, we report its performance as a reference. For its implementation, we first train U-Net on the labeled samples, and then employ the trained model to directly segment the unlabeled samples. Given $\Pi$-model, temporal ensembling and mean teacher are all current state-of-the-art models for semi-supervised learning, for our segmentation task, we simply modify them by replacing the softmax layers with a decoder as in \cite{yu2019uncertainty}. To evaluate these competitive baselines, we employ the Dice score (DSC in short) as the main evaluation metric, while precision, recall, and the Hausdorff distance as supplementary metrics. In our evaluation, higher values of DSC, precision and recall indicate better performances, whereas lower values of the Hausdorff distance are more preferable.

We use 80\% samples in the whole dataset as the training set, and the other 20\% as the test set. In the training set, we use 50\% as the labeled samples (\ie, 40\% samples of the whole dataset) and the other 50\% as the unlabeled samples. We report the comparison results in Table \ref{tab:sota_pancreas}. From this table, it is evident that our method outperforms these comparison methods in terms of DSC, recall and the Hausdorff distance, although our precision is worse than a few methods, \eg, temporal ensembling.
\begin{table}[htbp]
\footnotesize
\centering
\caption{The comparison on \re{2D} CT pancreas segmentation. We use 50\% of the training set as labeled samples and 50\% as unlabeled samples.}
\label{tab:sota_pancreas}
\renewcommand\arraystretch{1}
\begin{tabular}{c|cccc}
\toprule
Method & DSC (\%) & Prec. (\%) & Rec. (\%) & HD \scriptsize{(voxel)} \\
\midrule
\specialrule{0pt}{1pt}{1pt}
U-Net \cite{unet2015MICCAI} & 57.53 & 67.78 & 52.92 & 27.65\\
$\Pi$-model \cite{laine2016temporal} & 57.69 & 68.98 & 55.20 & 26.37\\
TE \cite{laine2016temporal} & 61.18 & \textbf{71.79} & 60.22 & 21.94\\
MT \cite{tarvainen2017mean} & 62.50 & 68.97 & 62.16 & 23.34\\
UA-MT \cite{yu2019uncertainty} & 63.82 & 63.47 & 73.12 & 20.64\\
\textbf{Ours} &\textbf{67.01} & 68.41 &\textbf{74.19} & \textbf{15.90}\\
\bottomrule
\end{tabular}
\end{table}

We also investigate the performance of our method with different proportions of the labeled samples in the training set. In particular, we introduce three different ratios of labeled samples to unlabeled samples, \ie, 1:4, 1:2, and 1:1. As is shown by their DSC values in Table \ref{tab:ratio_pancreas}, our method could consistently achieve better and more stable segmentation performance with different ratios of labeled to unlabeled samples than the comparison methods.
\begin{table}[htbp]
\footnotesize
\centering
\caption{The values of DSC (\%) with different ratios of labeled to unlabeled samples.}
\label{tab:ratio_pancreas}
\renewcommand\arraystretch{1}
\begin{tabular}{c|ccc}
\toprule
\multirow{2}*{Method} & \multicolumn{3}{c}{Labeled to unlabeled} \\
 & \textbf{1 : 4} & \textbf{1 : 2} & \textbf{1 : 1}\\
\midrule
\specialrule{0pt}{1pt}{1pt}
U-Net \cite{unet2015MICCAI} & 46.71 & 50.12 & 57.53 \\
$\Pi$-model \cite{laine2016temporal}  & 47.03 & 50.60 & 57.09 \\
Temporal Ensembling \cite{laine2016temporal} & 48.21 & 51.31 & 61.18 \\
Mean Teacher \cite{tarvainen2017mean} & 48.62 & 52.72 & 62.50 \\
UA-MT \cite{yu2019uncertainty} & 50.63 & 56.26 & 63.82 \\
\textbf{Ours} & \textbf{53.82} & \textbf{61.16} & \textbf{67.01} \\
\bottomrule
\end{tabular}
\end{table}

In addition, we also provide an ablation study to validate the efficacy of the proposed losses in this paper. Recall that we use main segmentation loss $\mathcal{L}_{crm}$ in Eqn. (\ref{eqn:crninitial}), two separate training losses, \ie, self-training in Eqn. (\ref{eqn:csnobjv}) and mean teacher loss in Eqn. (\ref{eqn:ucsnobjv}). To investigate if these proposed losses could benefit the final performance,
we introduce several variants of our method. In particular, we introduce six settings as below
\begin{itemize}
    \item \texttt{seg}: the model of merely using the main loss $\mathcal{L}_{crm}$ in Eqn. (\ref{eqn:crninitial}).
    \item \texttt{seg}+\texttt{st}: main segmentation loss and self-training loss. This means that we do not use uncertainty mask in CRM but directly employ self-training on unlabeled samples.
    \item \texttt{seg}+\texttt{st} (\texttt{mask}): main segmentation loss and self-training loss. Note that, this self-training loss is calculated only on certain region according to the mask in CRM, which indicates that the uncertain region is not involved in the training process.
    \item \texttt{seg}+\texttt{mt}: main segmentation loss and mean teacher loss. Here we do not use uncertainty mask in CRM, meaning that this setting directly employs mean teacher on the unlabeled samples.
    \item \texttt{seg}+\texttt{mt} (\texttt{mask}): main segmentation loss and mean teacher loss. Note that, the uncertain region is not involved in the training process.
    \item \texttt{ours}: our overall method, including the proposed CRM and separate self-training losses.
\end{itemize}
We report the results of DSC, precision, recall and HD in Table \ref{tab:abstu_pancreas}. We could observe that using all the components proposed in our work benefits the overall segmentation performance. Also, by comparing the setting of using and not using certain and uncertain masks, we validate the efficacy of certain mask predicted by our CRM.
\begin{table}[htbp]
\footnotesize
\centering
\caption{Ablation study on CT pancreas segmentation.}
\label{tab:abstu_pancreas}
\renewcommand\arraystretch{1}
\begin{tabular}{c|cccc}
\toprule
Method & DSC (\%) & Prec. (\%) & Rec. (\%) & HD \scriptsize{(voxel)} \\
\midrule
\specialrule{0pt}{1pt}{1pt}
\texttt{seg} & 60.10 & \textbf{69.67} & 54.68 & 25.07\\
\texttt{seg}+\texttt{st} & 62.33 & 68.72 & 63.07 & 22.59\\
\texttt{seg}+\texttt{st} (\texttt{mask}) & 64.23 & 67.90 & 71.99 & 19.53\\
\texttt{seg}+\texttt{mt} & 61.23 & 68.04 & 60.78 & 23.43\\
\texttt{seg}+\texttt{mt} (\texttt{mask}) & 63.69 & 69.60 & 70.23 & 20.40\\
\texttt{\textbf{ours}} & \textbf{67.01} & 68.41 & \textbf{74.19} & \textbf{15.90}\\
\bottomrule
\end{tabular}
\end{table}

\re{To further investigate the performance, we implement our method in its 3D version by utilizing V-Net and ResNet-18 as the backbone to compare with recent methods in Table~\ref{tab:sota_pancreas_3d}, including \textbf{DAN}~\cite{zhang2017deep}, \textbf{ADVNET}~\cite{vu2019advent}, \textbf{UA-MT}~\cite{yu2019uncertainty}, \textbf{SASSNet}~\cite{li2020shape}, \textbf{DTC}~\cite{luo2020semisupervised} and \textbf{UMCT} \cite{XIA2020101766}. In addition, the backbone of DAN, ADVNET, UA-MT, SASSNet and DTC is V-Net, whereas the backbone of UMCT is ResNet-18. Specifically, the backbone of V-Net consists of five layers with four downsampling and upsampling layers. The downsampling layer is implemented with 3D convolution with stride of 2 and the upsampling layer with 3D transposed convolution. Similar to the 2D setting discussed earlier, we first pretrain V-Net 30 epochs and then conduct self-training in the following 200 epochs. For the backbone of ResNet-18, we modify ResNet-18 into 3D version by extending the first 7$\times$7 convolution to 7$\times$7$\times$3 and changing all other 3$\times$3 convolutional layers into 3$\times$3$\times$1 that can be trained as a 3D convolutional layer. 
During the training stage, we only apply random crop as the augmentation method.
For fair comparison with these baselines, we utilize 12 labeled and 50 unlabeled volumes to train our model, as in all these baselines. The V-Net is only trained with 12 labeled data in the supervised setting, while other baseline methods are trained with both labeled and unlabeled data in the semi-supervised setting. In the inference phase, we adopt a sliding window strategy to obtain the final results with a stride of 16$\times$16$\times$4. We do not perform any postprocessing or model ensembling for fair comparison. 
Since all these methods follow the same setting, we directly report their best performance presented in their original literature. For evaluation metrics, we adopt DSC, Jaccard index, ASD, and HD, all widely used in the 3D setting. As observed from Table \ref{tab:sota_pancreas_3d}, 
our method outperforms all the baselines using V-Net as backbone. And our performance is better than UMCT (2v) and UMCT (6v), except UMCT (6v+). Note that, UMCT (6v+) utilizes multi-view model ensembling which has not been used in our method. Also, UMCT (6v+) is trained on an NVIDIA TITAN RTX GPU with 24GB memory for 24 hours \cite{XIA2020101766}, which is much more computationally expensive than ours (about 3 hours on an RTX 2080TI).}

\begin{table}[htbp]
\footnotesize
\centering
\caption{\re{The comparison on 3D CT pancreas segmentation.}}
\label{tab:sota_pancreas_3d}
\renewcommand\arraystretch{1}
\re{
\begin{tabular}{ccccc}
\toprule
Method & DSC (\%) & Jaccard (\%) & ASD \scriptsize{(voxel)} & HD \scriptsize{(voxel)} \\
\midrule
\specialrule{0pt}{1pt}{1pt}
\multicolumn{5}{c}{\textbf{Backbone: V-Net}} \\
\re{V-Net \cite{milletari2016v}}       & 69.96 & 55.55 & 1.64 & 14.27 \\
\re{DAN \cite{zhang2017deep}}         & 76.74 & 63.29 & 2.97 & 11.13 \\
\re{ADVNET \cite{vu2019advent}}      & 75.31 & 61.73 & 3.88 & 11.72 \\
\re{UA-MT \cite{yu2019uncertainty}}   & 77.26 & 63.82 & 3.06 & 11.90 \\
\re{SASSNet \cite{li2020shape}}       & 77.66 & 64.08 & 3.05 & 10.93 \\
\re{DTC \cite{luo2020semisupervised}} & 78.27 & 64.75 & 2.25 &  8.36  \\
\re{\textbf{Ours}}                             & \textbf{79.67} & \textbf{66.69} & \textbf{1.89} &  \textbf{7.59}  \\ 
\midrule
\multicolumn{5}{c}{\textbf{Backbone: ResNet-18}} \\
UMCT \scriptsize{(2v)} \cite{XIA2020101766}  & 79.77 & - & - & - \\
UMCT \scriptsize{(6v)} \cite{XIA2020101766}  & 80.35 & - & - & - \\
UMCT \scriptsize{(6v+)} \cite{XIA2020101766}  & \textbf{81.18} & - & - & - \\
\textbf{Ours}           & 80.58 & 67.91 & 2.27 & 8.34\\
\bottomrule
\end{tabular}
}
\end{table}

\re{Furthermore, we also validate the superiority of our uncertainty estimation method. Specifically, we replace the certain and uncertain mask generation module with other uncertainty estimation methods, including test augmentation~\cite{Marcunc20}, MC dropout~\cite{gal16unc} and aleatoric uncertainty~\cite{kendall2017uncertainties}. For the test augmentation, we only employ random crop in the 3D setting, and apply random horizontal flip and random scale with ratio between $[0.8, 1.1]$ in the 2D setting, respectively. We perform forward pass 20 times to obtain the average result. For MC dropout, we add the dropout to the last downsampled feature and the final feature with a dropout rate of 0.5. Then, we conduct forward pass 50 times with dropout enabled in the test time and average their results. For both test augmentation and MC dropout, we adopt the confidence threshold of 0.8 for the certain and uncertain mask generation. The aleatoric uncertainty captures the inherent  uncertainty in the observation and assumes the normal distribution in logits. We implement this method with two branches of the same structure as our separate self-training branches and output the $f$ and $\sigma$. The noise is sampled 20 times in the 2D setting and 10 times in the 3D setting, respectively. To generate the certain and uncertain masks, we manually set the standard deviation less than 0.001 as the certain region and the rest as the uncertain region. The result in Table \ref{tab:uncertainty_evaluation} shows the advantage of our method in both 2D and 3D.}

\begin{table}[htbp]
\footnotesize
\centering
\caption{\re{The comparison on DSC of different uncertainty estimation methods on both 2D and 3D CT pancreas segmentation. In the 2D setting, we employ 50\% labeled data, while in the 3D setting, we utilize 20\% labeled data, which is consistent with our original setting.}}
\label{tab:uncertainty_evaluation}
\renewcommand\arraystretch{1}
\re{
\begin{tabular}{c|cc}
\toprule
Method & 2D setting & 3D setting \\
\midrule 
\re{Test Augmentation \cite{Marcunc20}} & 62.75 \% & 78.27 \% \\
\re{MC Dropout \cite{gal16unc}}  & 63.03 \% & 77.07 \% \\
\re{Aleatoric Uncertainty \cite{kendall2017uncertainties}} & 64.91 \% & 78.67 \% \\
\textbf{Ours} & \textbf{67.01 \%} & \textbf{79.67 \%}\\
\specialrule{0pt}{1pt}{1pt}
\bottomrule
\end{tabular}
}
\end{table}

\begin{figure}[htbp]
\centering
\subfigure{
\begin{minipage}[t]{0.13\linewidth}
\centering
\includegraphics[width=0.53in]{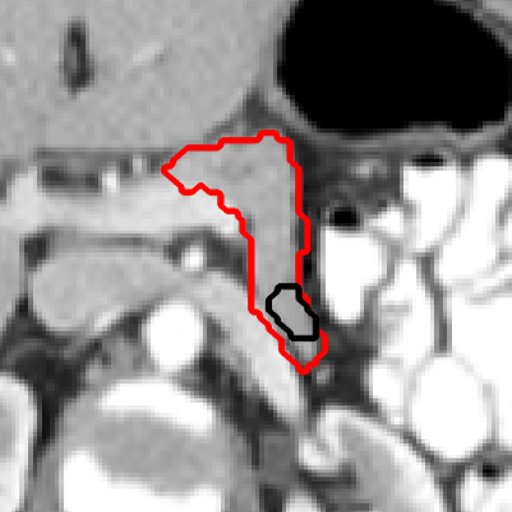}
\end{minipage}}
\subfigure{
\begin{minipage}[t]{0.13\linewidth}
\centering
\includegraphics[width=0.53in]{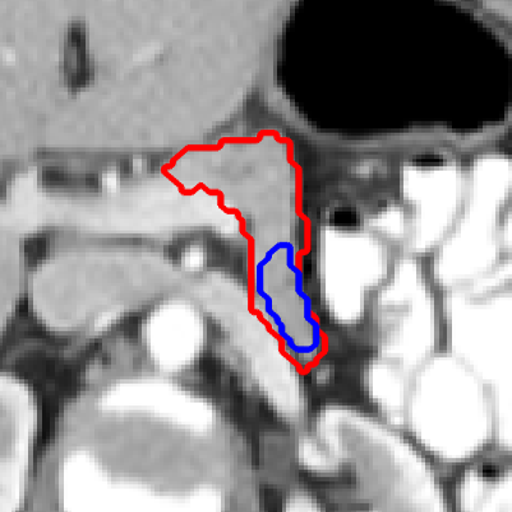}
\end{minipage}}
\subfigure{
\begin{minipage}[t]{0.13\linewidth}
\centering
\includegraphics[width=0.53in]{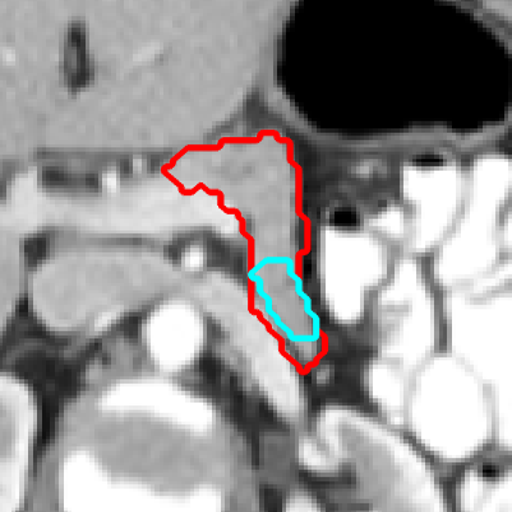}
\end{minipage}}
\subfigure{
\begin{minipage}[t]{0.13\linewidth}
\centering
\includegraphics[width=0.53in]{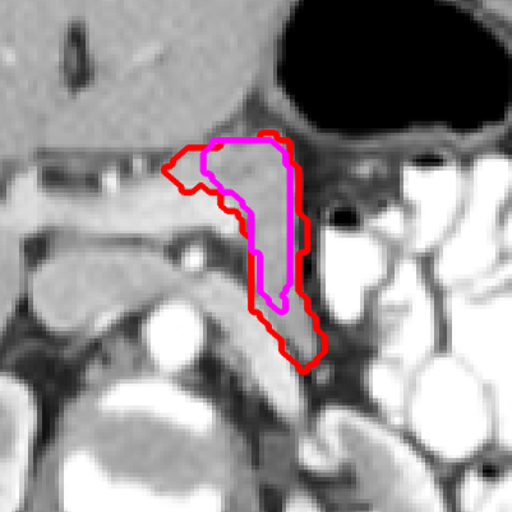}
\end{minipage}}
\subfigure{
\begin{minipage}[t]{0.13\linewidth}
\centering
\includegraphics[width=0.53in]{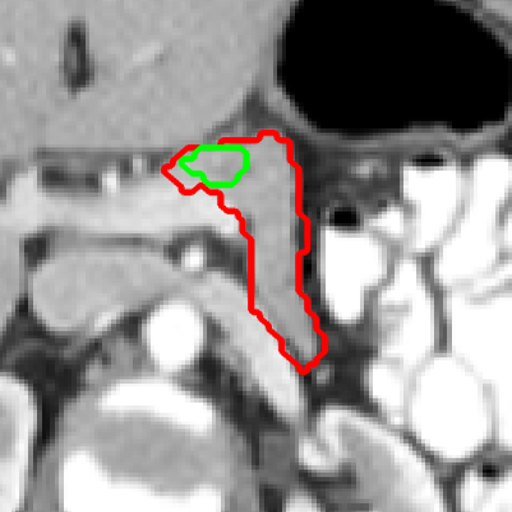}
\end{minipage}}
\subfigure{
\begin{minipage}[t]{0.13\linewidth}
\centering
\includegraphics[width=0.53in]{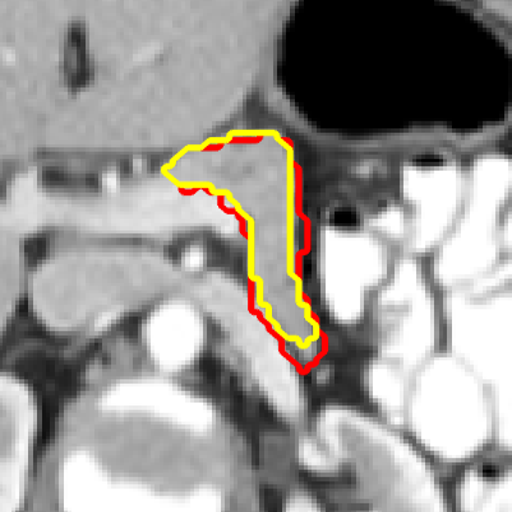}
\end{minipage}}\\
\vspace*{-0.25cm}
\subfigure{
\begin{minipage}[t]{0.13\linewidth}
\centering
\includegraphics[width=0.53in]{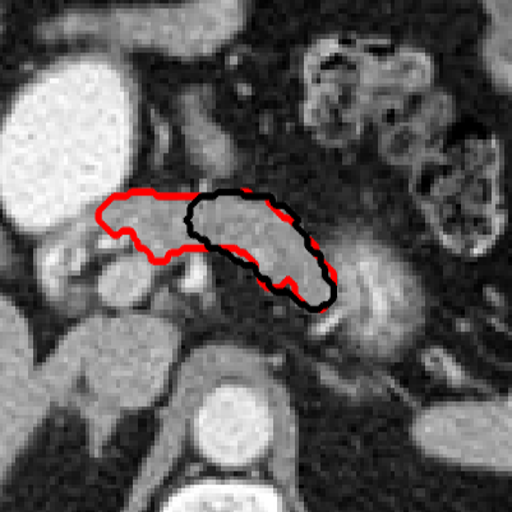}
\end{minipage}}
\subfigure{
\begin{minipage}[t]{0.13\linewidth}
\centering
\includegraphics[width=0.53in]{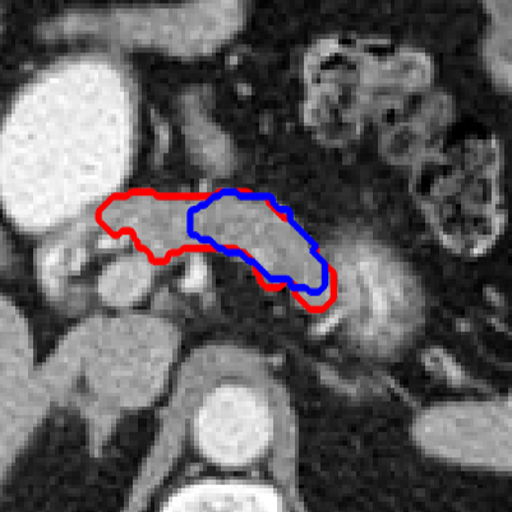}
\end{minipage}}
\subfigure{
\begin{minipage}[t]{0.13\linewidth}
\centering
\includegraphics[width=0.53in]{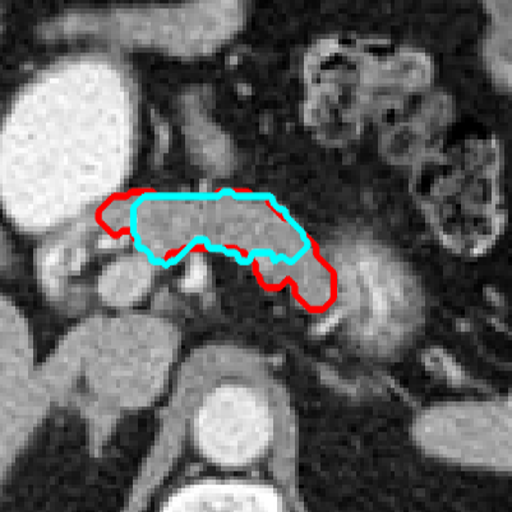}
\end{minipage}}
\subfigure{
\begin{minipage}[t]{0.13\linewidth}
\centering
\includegraphics[width=0.53in]{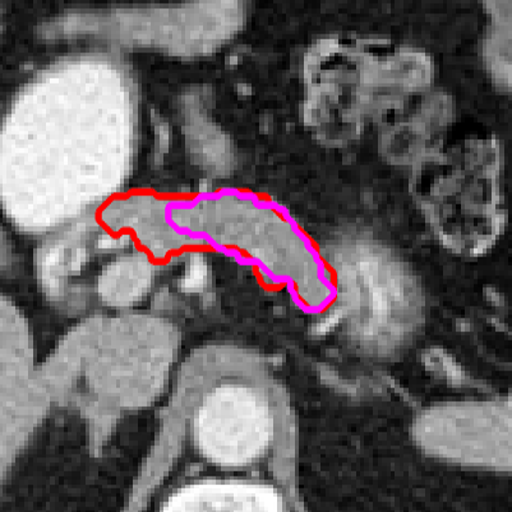}
\end{minipage}}
\subfigure{
\begin{minipage}[t]{0.13\linewidth}
\centering
\includegraphics[width=0.53in]{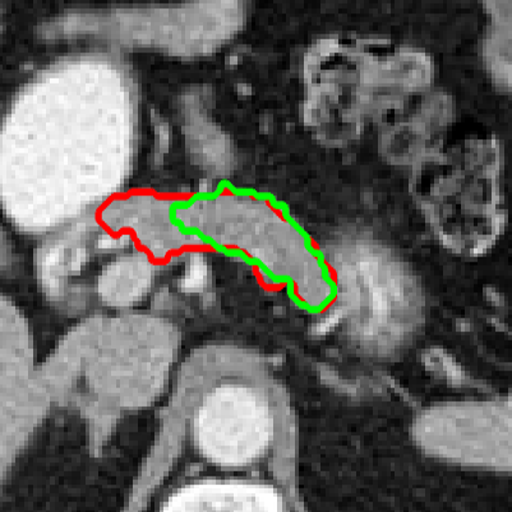}
\end{minipage}}
\subfigure{
\begin{minipage}[t]{0.13\linewidth}
\centering
\includegraphics[width=0.53in]{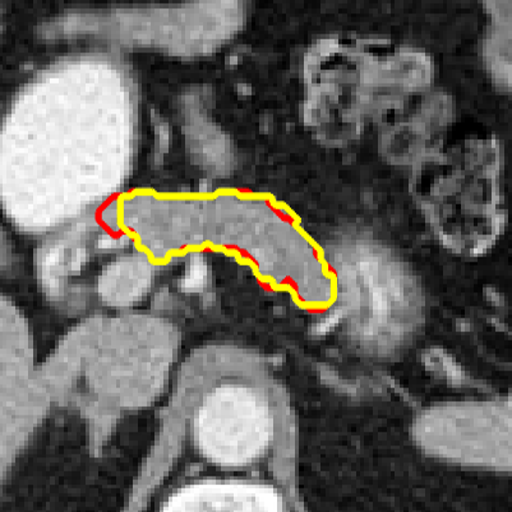}
\end{minipage}}\\
\vspace*{-0.25cm}
\subfigure{
\begin{minipage}[t]{0.13\linewidth}
\centering
\includegraphics[width=0.53in]{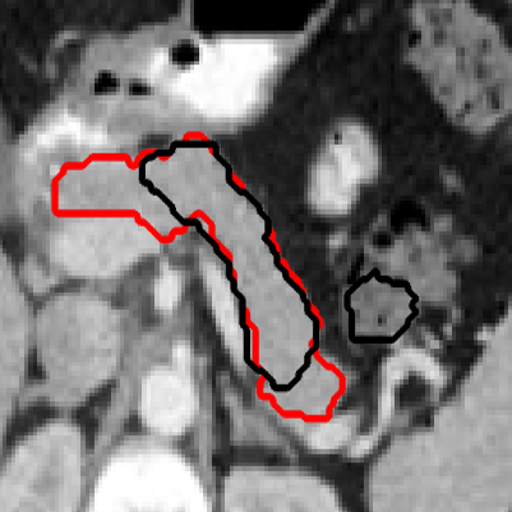}
\centerline{\scriptsize{MT}}
\end{minipage}}
\subfigure{
\begin{minipage}[t]{0.13\linewidth}
\centering
\includegraphics[width=0.53in]{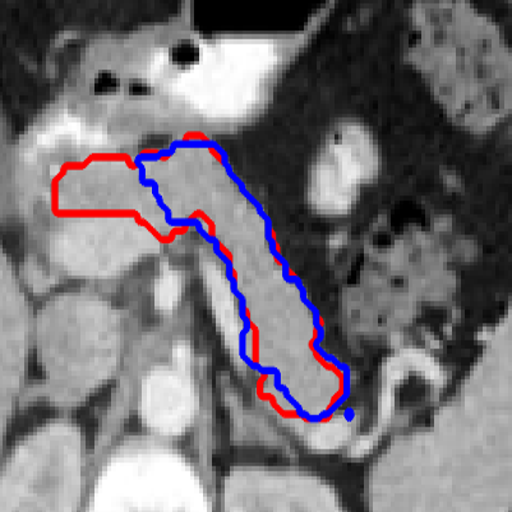}
\centerline{\scriptsize{$\Pi$-model}}
\end{minipage}}
\subfigure{
\begin{minipage}[t]{0.13\linewidth}
\centering
\includegraphics[width=0.53in]{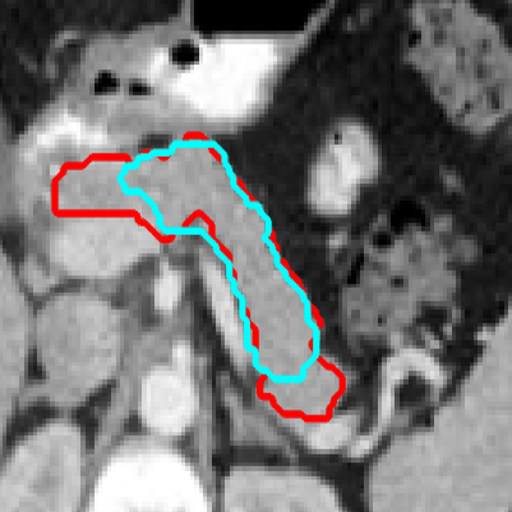}
\centerline{\scriptsize{TE}}
\end{minipage}}
\subfigure{
\begin{minipage}[t]{0.13\linewidth}
\centering
\includegraphics[width=0.53in]{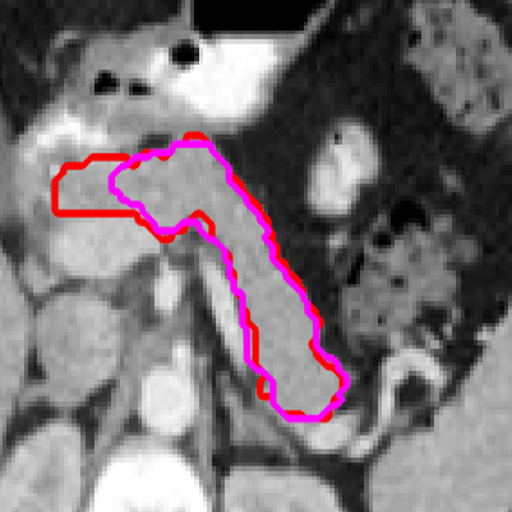}
\centerline{\scriptsize{U-Net}}
\end{minipage}}
\subfigure{
\begin{minipage}[t]{0.13\linewidth}
\centering
\includegraphics[width=0.53in]{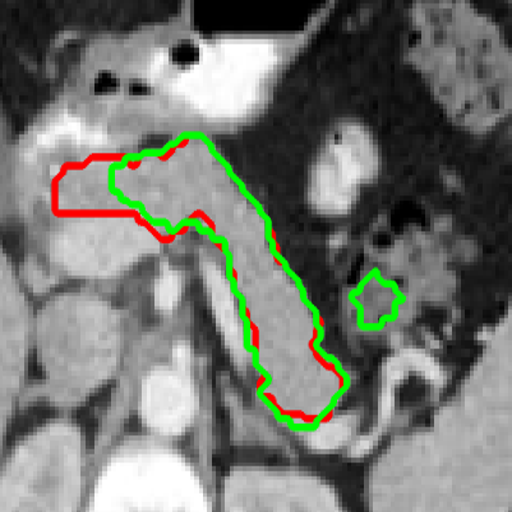}
\centerline{\scriptsize{UA-MT}}
\end{minipage}}
\subfigure{
\begin{minipage}[t]{0.13\linewidth}
\centering
\includegraphics[width=0.53in]{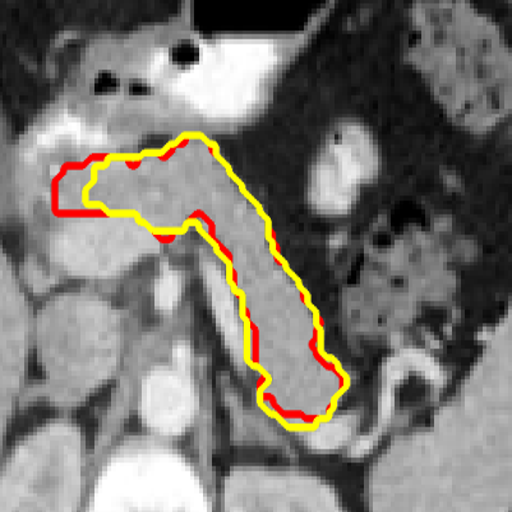}
\centerline{\scriptsize{Ours}}
\end{minipage}}
\caption{Visual segmentation examples in the \textit{pancreas} dataset with each row showing one CT scan. The red, black, blue, cyan, magenta, green and yellow curves denote the corresponding results of ground-truth, mean teacher \cite{tarvainen2017mean}, $\Pi$-model \cite{laine2016temporal}, temporal ensembling \cite{laine2016temporal}, U-Net \cite{unet2015MICCAI}, UA-MT \cite{yu2019uncertainty} and ours, respectively.}
\label{fig:exp:visualpancreas}
\end{figure}

Finally, we illustrate several segmentation examples of mean teacher, $\Pi$-model, temporal ensembling, U-Net, UA-MT and ours in Figure \ref{fig:exp:visualpancreas}. The red curves indicate the ground truth while the segmentation results of respective methods are indicated in different colors. These examples show that the segmentation results of our method are very close to the ground truth. In addition, our method also shows its advantage in segmentation on several difficult cases.

\begin{figure}[htbp]
\centering
\subfigure{
\begin{minipage}[t]{0.13\linewidth}
\centering
\includegraphics[width=0.53in]{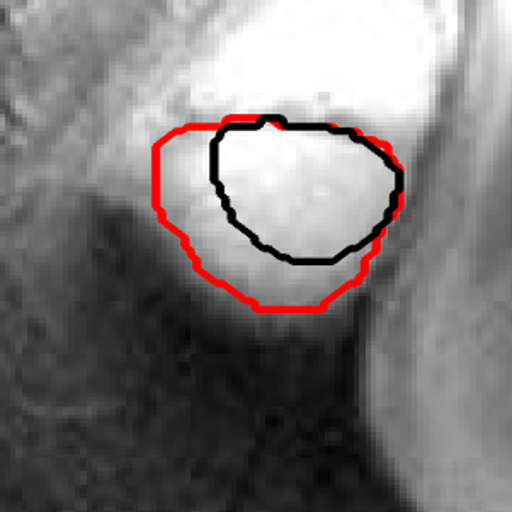}
\end{minipage}}
\subfigure{
\begin{minipage}[t]{0.13\linewidth}
\centering
\includegraphics[width=0.53in]{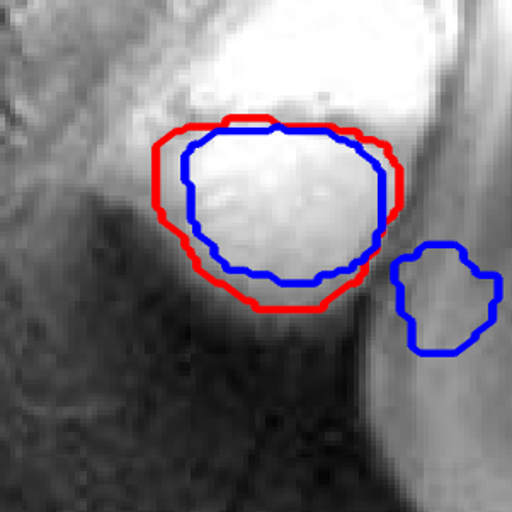}
\end{minipage}}
\subfigure{
\begin{minipage}[t]{0.13\linewidth}
\centering
\includegraphics[width=0.53in]{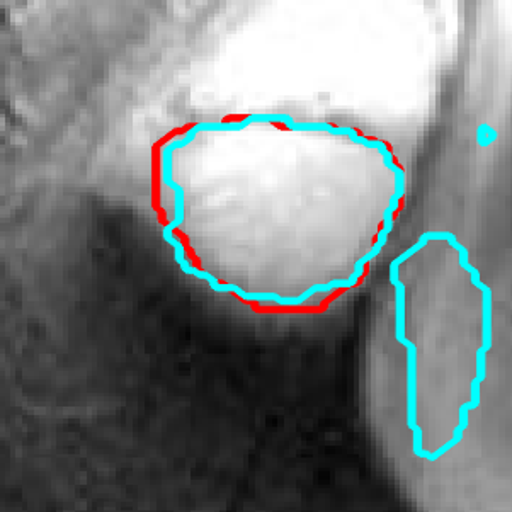}
\end{minipage}}
\subfigure{
\begin{minipage}[t]{0.13\linewidth}
\centering
\includegraphics[width=0.53in]{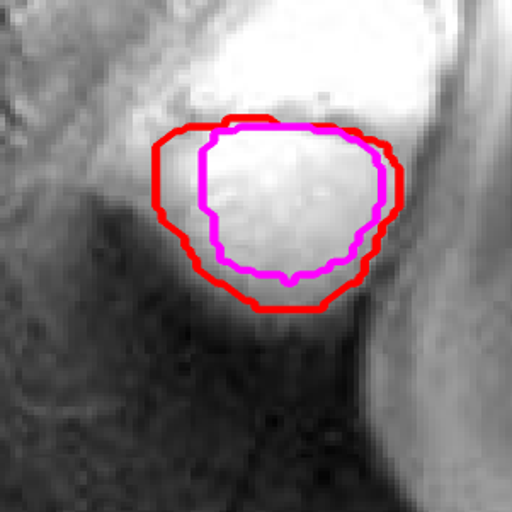}
\end{minipage}}
\subfigure{
\begin{minipage}[t]{0.13\linewidth}
\centering
\includegraphics[width=0.53in]{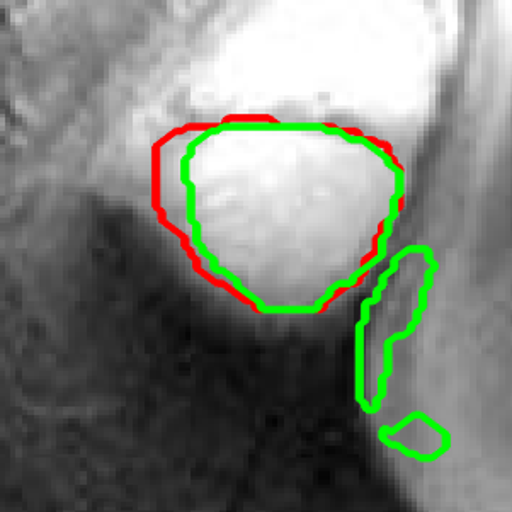}
\end{minipage}}
\subfigure{
\begin{minipage}[t]{0.13\linewidth}
\centering
\includegraphics[width=0.53in]{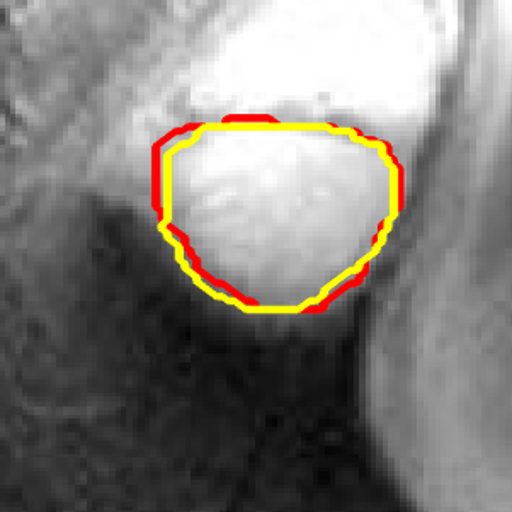}
\end{minipage}}\\
\vspace*{-0.25cm}
\subfigure{
\begin{minipage}[t]{0.13\linewidth}
\centering
\includegraphics[width=0.53in]{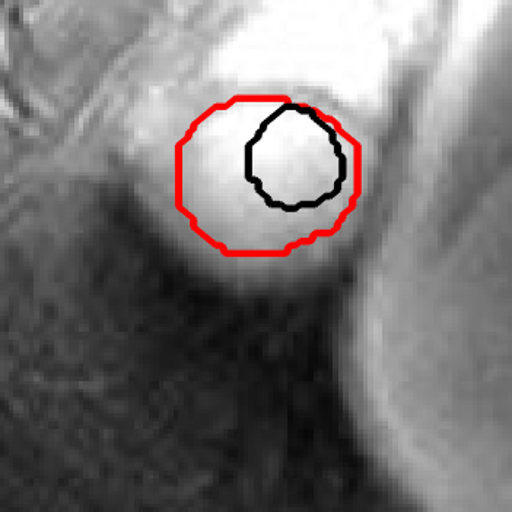}
\end{minipage}}
\subfigure{
\begin{minipage}[t]{0.13\linewidth}
\centering
\includegraphics[width=0.53in]{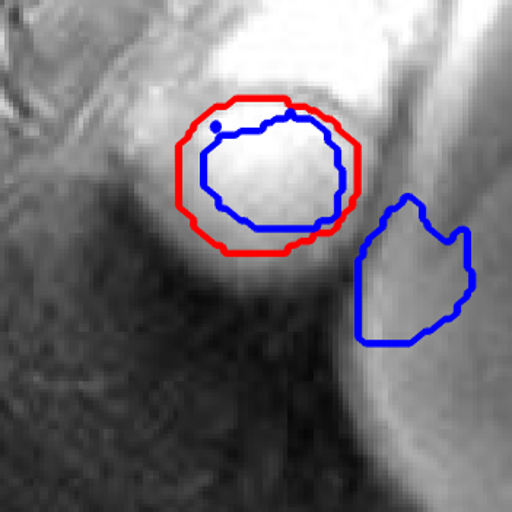}
\end{minipage}}
\subfigure{
\begin{minipage}[t]{0.13\linewidth}
\centering
\includegraphics[width=0.53in]{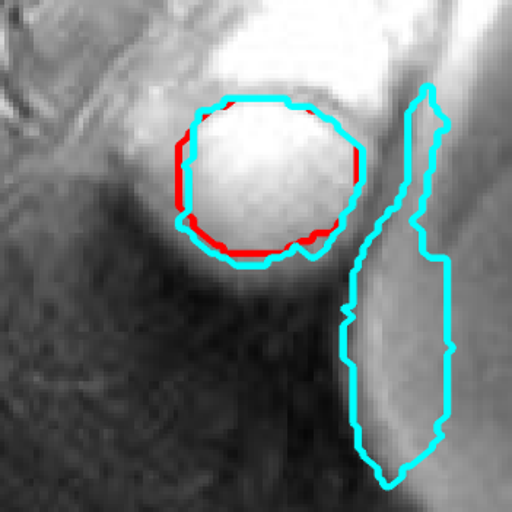}
\end{minipage}}
\subfigure{
\begin{minipage}[t]{0.13\linewidth}
\centering
\includegraphics[width=0.53in]{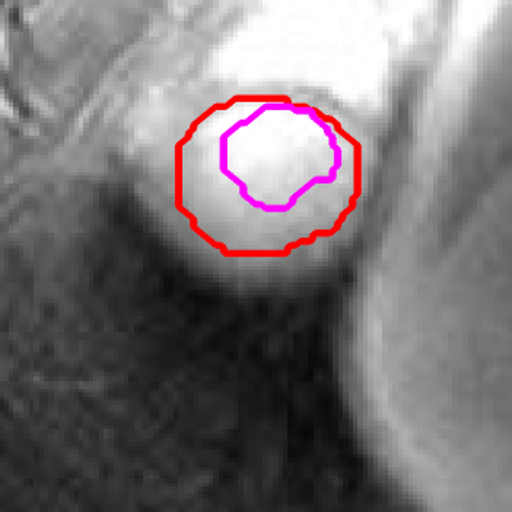}
\end{minipage}}
\subfigure{
\begin{minipage}[t]{0.13\linewidth}
\centering
\includegraphics[width=0.53in]{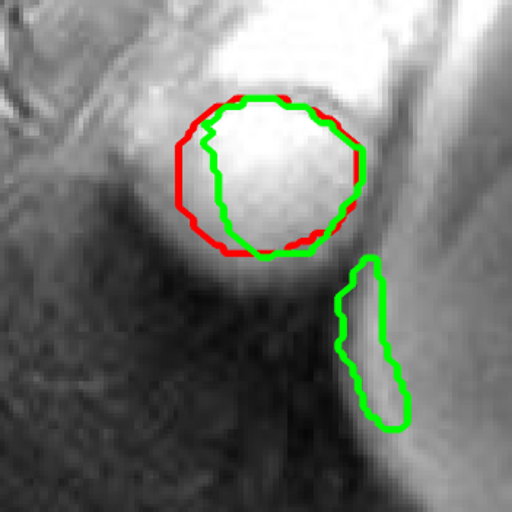}
\end{minipage}}
\subfigure{
\begin{minipage}[t]{0.13\linewidth}
\centering
\includegraphics[width=0.53in]{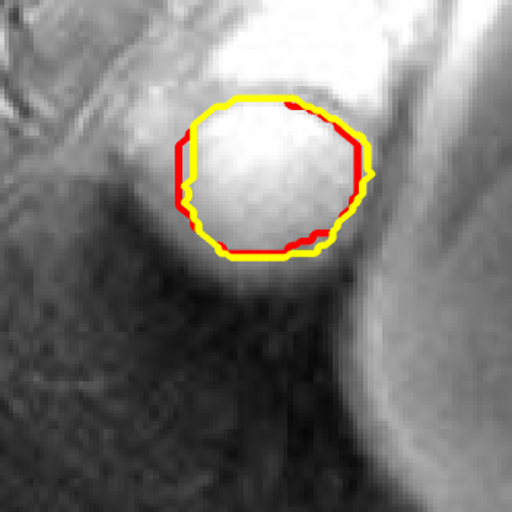}
\end{minipage}}\\
\vspace*{-0.25cm}
\subfigure{
\begin{minipage}[t]{0.13\linewidth}
\centering
\includegraphics[width=0.53in]{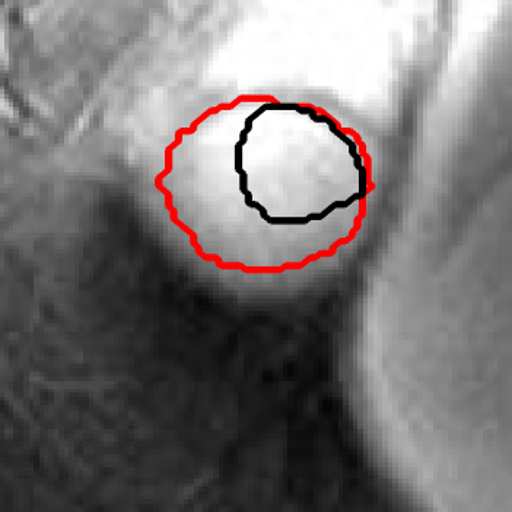}
\centerline{\scriptsize{MT}}
\end{minipage}}
\subfigure{
\begin{minipage}[t]{0.13\linewidth}
\centering
\includegraphics[width=0.53in]{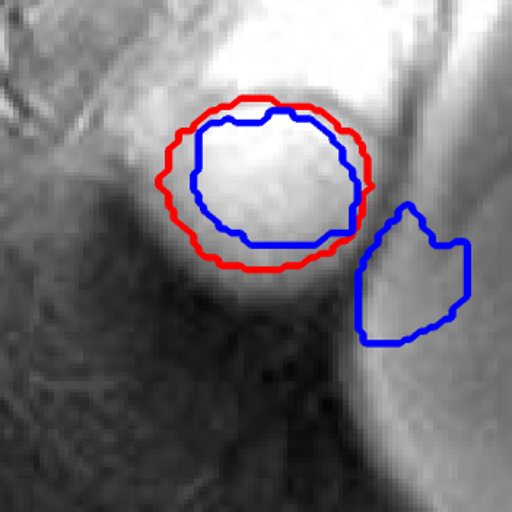}
\centerline{\scriptsize{$\Pi$-model}}
\end{minipage}}
\subfigure{
\begin{minipage}[t]{0.13\linewidth}
\centering
\includegraphics[width=0.53in]{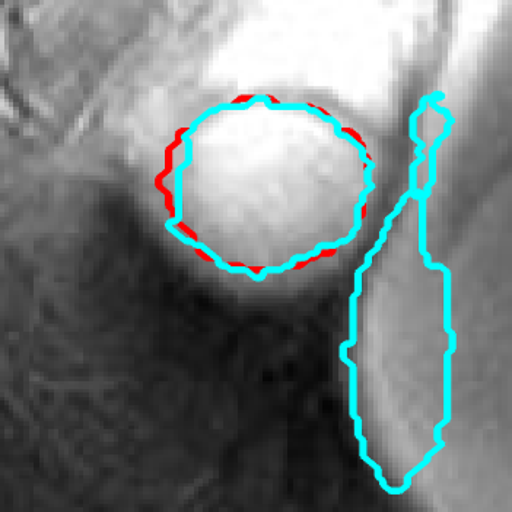}
\centerline{\scriptsize{TE}}
\end{minipage}}
\subfigure{
\begin{minipage}[t]{0.13\linewidth}
\centering
\includegraphics[width=0.53in]{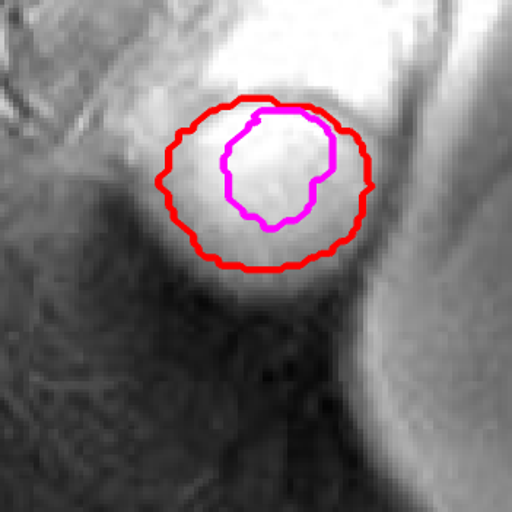}
\centerline{\scriptsize{U-Net}}
\end{minipage}}
\subfigure{
\begin{minipage}[t]{0.13\linewidth}
\centering
\includegraphics[width=0.53in]{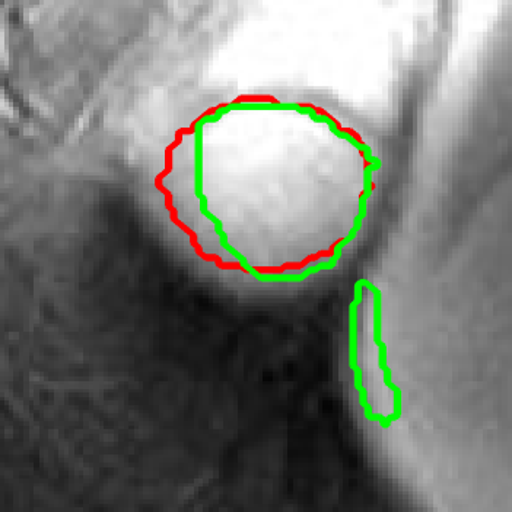}
\centerline{\scriptsize{UA-MT}}
\end{minipage}}
\subfigure{
\begin{minipage}[t]{0.13\linewidth}
\centering
\includegraphics[width=0.53in]{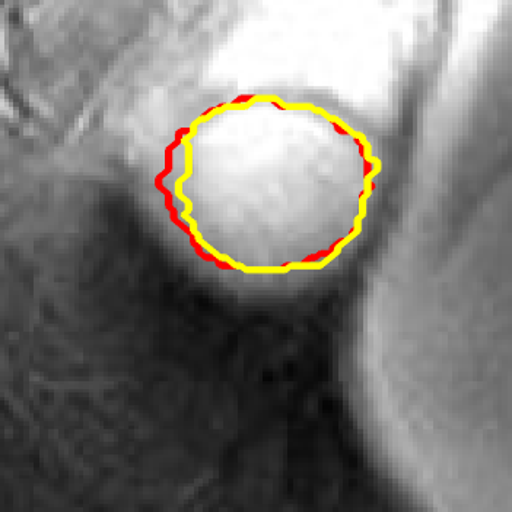}
\centerline{\scriptsize{Ours}}
\end{minipage}}
\caption{Visual segmentation examples in the \textit{endocardium} dataset. The red, black, blue, cyan, magenta, green and yellow curves denote the corresponding result of ground-truth, mean teacher \cite{tarvainen2017mean}, $\Pi$-model \cite{laine2016temporal}, temporal ensembling \cite{laine2016temporal}, U-Net \cite{unet2015MICCAI}, UA-MT \cite{yu2019uncertainty} and ours, respectively.}
\label{fig:exp:visualcardiac}
\end{figure}

\subsection{Results on MR Endocardium}
As in the baselines in CT pancreas segmentation, we also utilize \textbf{U-Net}, \textbf{$\Pi$-model}, \textbf{temporal ensembling}, \textbf{mean teacher} and \textbf{UA-MT} as the comparison methods in MR endocardium segmentation, and use DSC, precision, recall and the Hausdorff distance as our evaluation metrics. 
\re{The implementation of these baseline methods in MR endocardium are the same as that in CT pancreas. Also, note that we only conduct the 2D setting since the 3D information on MR endocardium dataset is currently unavailable on their website.}
By setting the ratio of labeled to unlabeled samples as 1:4, we obtain the results in Table \ref{tab:sotacardiac}. We notice that our method outperforms these comparison methods again, especially in DSC, on which our method outperforms by a large margin.
\begin{table}[htbp]
\footnotesize
\centering
\caption{Comparison of different methods on the MR endocardium dataset.}
\label{tab:sotacardiac}
\renewcommand\arraystretch{1}
\begin{tabular}{c|cccc}
\toprule
Method & DSC (\%) & Prec. (\%) & Rec. (\%) & HD \scriptsize{(voxel)} \\
\midrule
U-Net\cite{unet2015MICCAI}  & 76.13 & 82.73 & 74.62 & 15.80\\
$\Pi$-model \cite{laine2016temporal} & 77.20 & 80.11 & 89.14 & 16.15\\
TE \cite{laine2016temporal} & 78.09 & 81.25 & 79.68 & 14.61\\
MT \cite{tarvainen2017mean} & 79.60 & 82.98 & 79.33 & 16.92\\
UA-MT \cite{yu2019uncertainty} & 82.02 & \textbf{83.39} & 92.21 & 13.70\\
\textbf{Ours} & \textbf{86.67} & 83.21 & \textbf{93.19} & \textbf{10.91}\\
\bottomrule
\end{tabular}
\end{table}

We illustrate several examples of segmentation results in Figure \ref{fig:exp:visualcardiac}. Similarly, the red, black, blue, cyan, magenta, green and yellow curves denote the corresponding result of ground-truth, mean teacher, $\Pi$-model, temporal ensembling, U-Net, UA-MT and ours, respectively. As shown in Figure \ref{fig:exp:visualcardiac}, the results of our method are more accurate than those of others.

\begin{figure}[htbp]
\centering
\subfigure{
\begin{minipage}[t]{0.13\linewidth}
\centering
\includegraphics[width=0.53in]{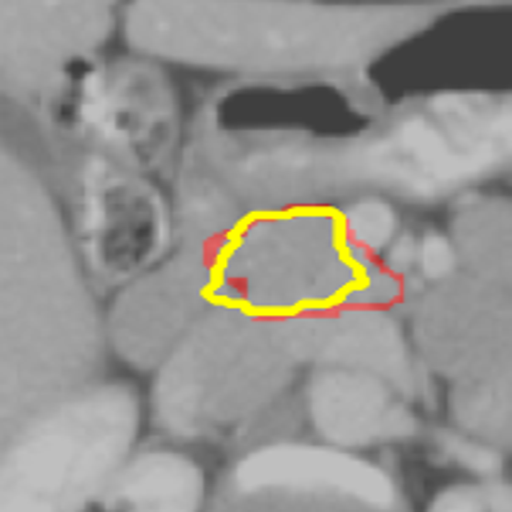}
\end{minipage}}
\subfigure{
\begin{minipage}[t]{0.13\linewidth}
\centering
\includegraphics[width=0.53in]{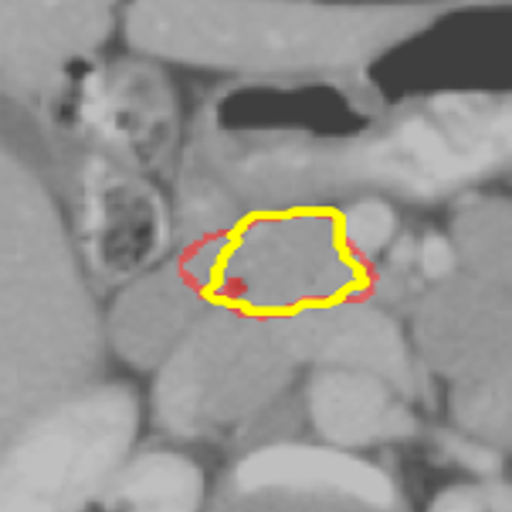}
\end{minipage}}
\subfigure{
\begin{minipage}[t]{0.13\linewidth}
\centering
\includegraphics[width=0.53in]{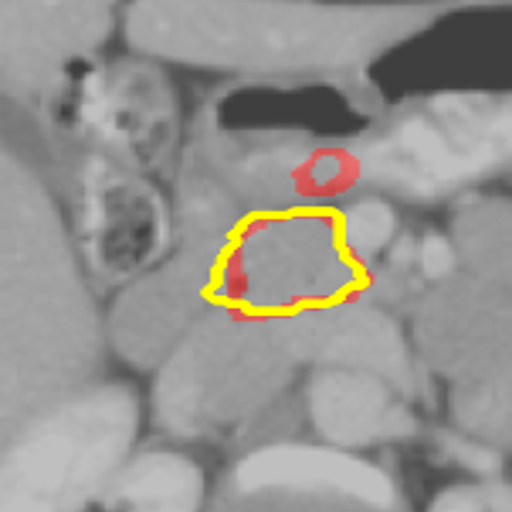}
\end{minipage}}
\subfigure{
\begin{minipage}[t]{0.13\linewidth}
\centering
\includegraphics[width=0.53in]{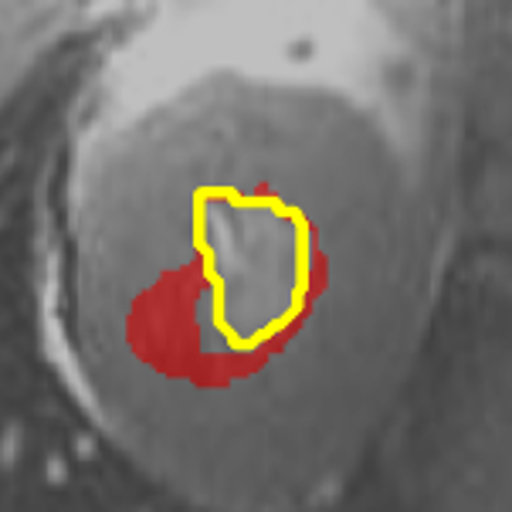}
\end{minipage}}
\subfigure{
\begin{minipage}[t]{0.13\linewidth}
\centering
\includegraphics[width=0.53in]{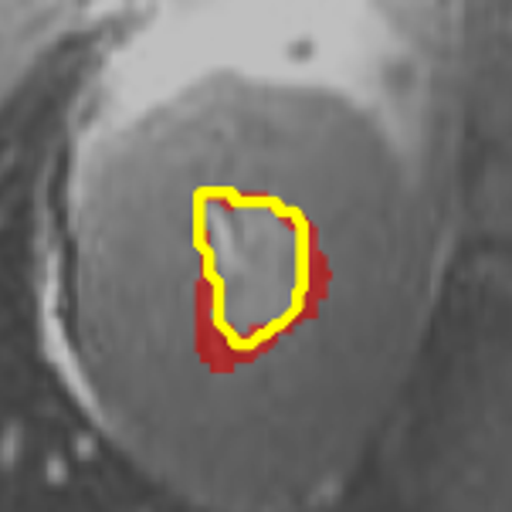}
\end{minipage}}
\subfigure{
\begin{minipage}[t]{0.13\linewidth}
\centering
\includegraphics[width=0.53in]{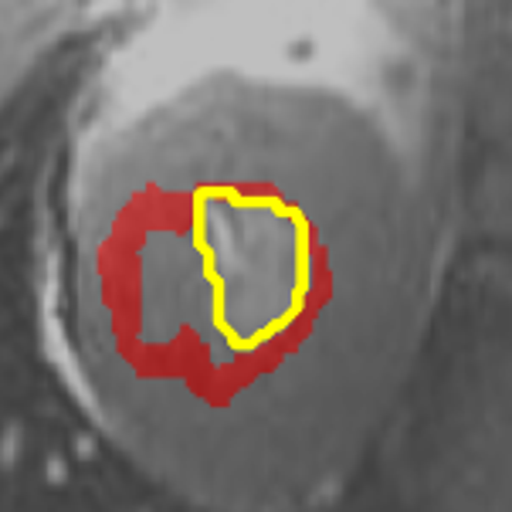}
\end{minipage}}\\
\vspace*{-0.25cm}
\subfigure{
\begin{minipage}[t]{0.13\linewidth}
\centering
\includegraphics[width=0.53in]{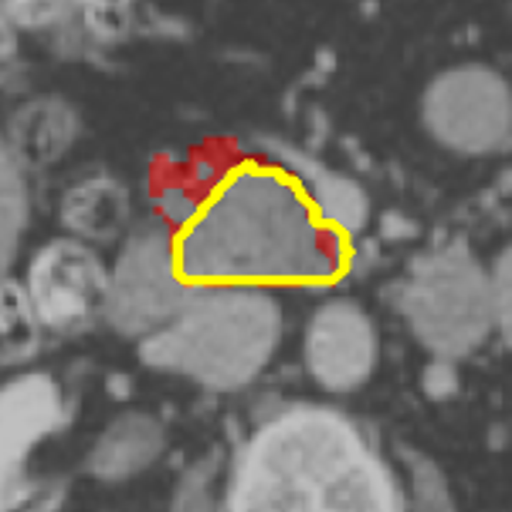}
\centerline{\scriptsize{$\alpha$=2}}
\end{minipage}}
\subfigure{
\begin{minipage}[t]{0.13\linewidth}
\centering
\includegraphics[width=0.53in]{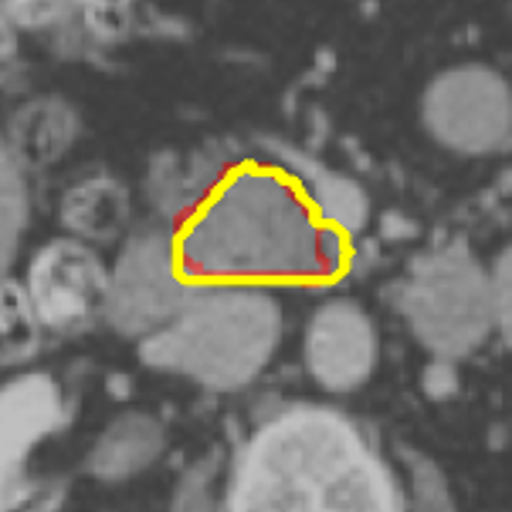}
\centerline{\scriptsize{$\alpha$=5}}
\end{minipage}}
\subfigure{
\begin{minipage}[t]{0.13\linewidth}
\centering
\includegraphics[width=0.53in]{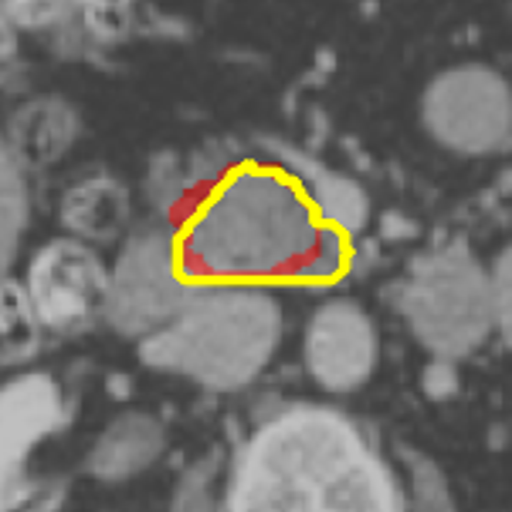}
\centerline{\scriptsize{$\alpha$=10}}
\end{minipage}}
\subfigure{
\begin{minipage}[t]{0.13\linewidth}
\centering
\includegraphics[width=0.53in]{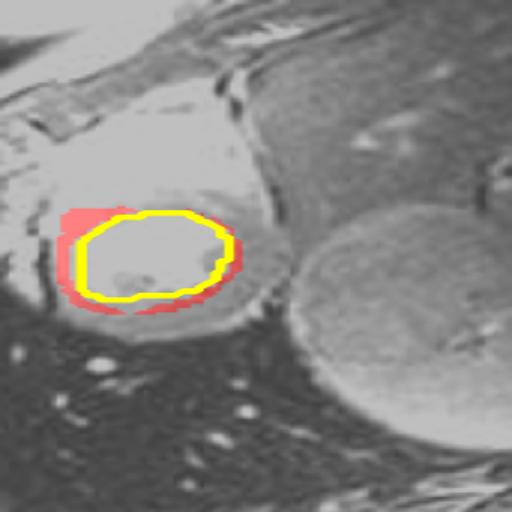}
\centerline{\scriptsize{$\alpha$=2}}
\end{minipage}}
\subfigure{
\begin{minipage}[t]{0.13\linewidth}
\centering
\includegraphics[width=0.53in]{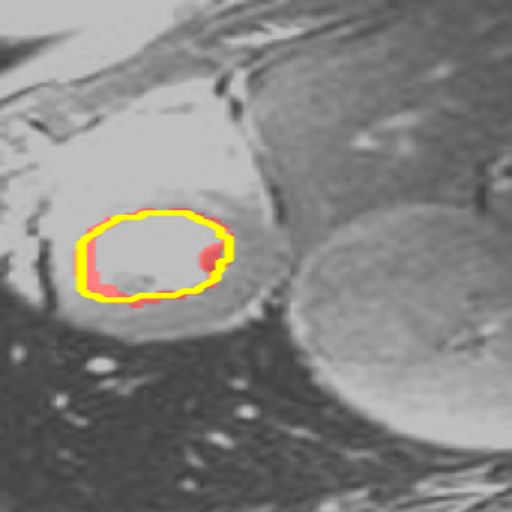}
\centerline{\scriptsize{$\alpha$=5}}
\end{minipage}}
\subfigure{
\begin{minipage}[t]{0.13\linewidth}
\centering
\includegraphics[width=0.53in]{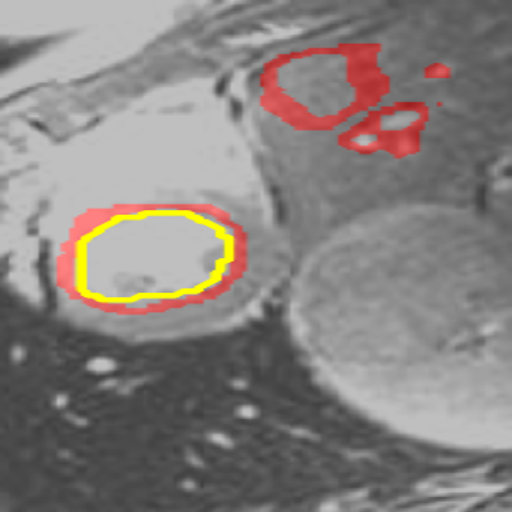}
\centerline{\scriptsize{$\alpha$=10}}
\end{minipage}}
\caption{The initial uncertainty masks by setting $\alpha$ as 2, 5 and 10, on the \textit{pancreas} and \textit{endocardium} datasets, respectively. The yellow curves indicate the ground truth and the red indicate the uncertain segmentation.}\label{fig:visual_costdiffpancendo}
\end{figure}

To better understand the setting of misclassification costs in our CRM, we conduct an experiment to investigate the performance of our method with different misclassification costs (\ie, $\alpha$ in Eqn. (\ref{eqn:crninitial})). Specifically, we set $\alpha$ to 2, 5, and 10, respectively, to check the change of performance on CT pancreas and MR endocardium segmentation. We report the results in Table \ref{tab:abstucosts}, which demonstrates that our segmentation results are relatively stable and superior to the baseline. \re{In this table, $\alpha$=1 ($\star$) indicates the method using different initializations for two decoders $\mathbf{D}_{rad}$ and $\mathbf{D}_{con}$ but with equal value of misclassification cost $\alpha$. In this setting, these two decoders first generate different outputs, and then, we calculate their difference as the uncertain mask. This setting is widely used in previous studies as another way of uncertainty estimation.} Moreover, we provide some visual examples of initial uncertainty masks with different values of $\alpha$ on CT pancreas and MR endocardium datasets in Figure \ref{fig:visual_costdiffpancendo}. 
We find that setting $\alpha$ to 5 is the best, since values too large or too small might introduce some noise during segmentation.
\begin{table}[htbp]
\footnotesize
\centering
\caption{The segmentation accuracy on pancreas and endocardium datasets with different values of $\alpha$.}
\label{tab:abstucosts}
\renewcommand\arraystretch{1}
\begin{tabular}{cc|cccc}
\toprule
Dataset & $\alpha$ & DSC (\%) & Prec. (\%) & Rec. (\%) & HD \scriptsize{(voxel)}
\\
\midrule
\multirow{3}*{\scriptsize{Pancreas}}  
 & \re{$\alpha$=1 ($\star$)} & \re{61.83} & \re{63.44} & \re{70.03} & \re{17.34} \\
 & $\alpha$=2 & 64.80 & 68.18 & 69.69 & 19.85 \\
 & $\alpha$=5 & \textbf{67.01} & \textbf{68.41} & \textbf{74.19} & \textbf{15.90} \\
 & $\alpha$=10 & 62.56 & 67.47 & 61.73 & 24.62 \\
\specialrule{0pt}{1pt}{1pt}
\midrule
\specialrule{0pt}{1pt}{1pt}
\multirow{3}*{\scriptsize{Endocardium}}  & \re{$\alpha$=1 ($\star$)} & \re{83.40} & \re{82.73} & \re{92.39} & \re{12.58} \\
& $\alpha$=2 & 85.42 & 78.71 & 97.01 & 12.19 \\
 & $\alpha$=5 & \textbf{86.67} & \textbf{83.21} & 93.19 & \textbf{10.91}\\
 & $\alpha$=10 & 83.62 & 76.45 & \textbf{97.15} & 31.36\\
\bottomrule
\end{tabular}
\end{table}

\subsection{ACDC Segmentation}
To investigate if our method performs well on multi-class segmentation tasks, we further evaluate our method and other baselines on the ACDC segmentation dataset.
For the baselines, we introduce several semi-supervised segmentation models, including: \textbf{pseudo label} (\textbf{PL}) \cite{lee2013pseudo}, \textbf{VAT} \cite{miyato2018virtual}, \textbf{mean teacher} (\textbf{MT}) \cite{tarvainen2017mean}, \textbf{DCT-Seg} \cite{peng2020deep}, \re{\textbf{ADVNET} \cite{vu2019advent}, \textbf{Mut. Info.} \cite{peng2020mutual}, \textbf{Cons. Reg.} \cite{peng2020mutual}, \textbf{MSE} \cite{peng2020mutual}, \textbf{KL} \cite{peng2020mutual} and \textbf{Peng \etal’s method} \cite{2021Boosting}}. We also list the fully-supervised setting as a reference for the upper bound of semi-supervised segmentation methods. 

Specifically, for the multi-class segmentation, we modify our CoraNet by introducing several binary segmentation sub-tasks. To integrate the results of these sub-tasks, we ensemble their outputs by majority voting with their confidence. We use the same data augmentation strategy as that in DCT-Seg \cite{peng2020deep} that includes random rotation, flip, and random crop of 85-95\% area on the original image. 

Our goal is to segment three classes: right ventricle (\textbf{RV}), left ventricle (\textbf{LV}) cavities, and the myocardium (epicardial contour more specifically, or \textbf{Myo} in short). Then, the average segmentation accuracy of these three classes is reported. 
To evaluate the performance, we introduce the Dice score (DSC) as the evaluation metric. \re{For fair comparison with baselines,} we use 20\% samples in the dataset as the labeled samples, and the rest as the unlabeled samples, as consistent with the setting introduced in \cite{peng2020deep} and \cite{peng2020mutual}. 


We report the values of DSC of our method and the comparison methods in Table \ref{tab:sota_acdc}. \re{Since all these methods are evaluated under the same setting, we directly report their performance presented in their literature. Note that, since the number of slices along z-axis is sometimes limited (\ie, less than 10), thus making 3D convolutions hard to perform, above methods usually adopted the 2D setting \cite{vu2019advent}\cite{peng2020mutual}\cite{2021Boosting}. For fair comparison, we follow their 2D setting.} As shown in this table, not only has our method largely outperformed the comparison methods \re{on the mean segmentation accuracy}, our results are also close to those of the fully-supervised trained model.
\begin{table}
\scriptsize
\centering
\caption{The comparison of different methods on the ACDC dataset. Reported values are averages (standard deviation in parentheses) for 3 runs with different random seeds.}
\label{tab:sota_acdc}
\renewcommand\arraystretch{1.1}
\begin{tabular}{c|cccc}
\toprule
\multirow{2}*{Method} & \multicolumn{4}{c}{DSC (\%)} \\
 ~ & RV & Myo & LV & mean\\
\midrule
PL \cite{lee2013pseudo} & 74.60 (0.32) & 78.91 (0.21) & 85.79 (0.17) & 79.77 (0.14)\\
VAT \cite{miyato2018virtual} & 72.78 (0.39) & 80.81 (0.21) & 87.60 (0.18) & 80.39 (0.15)\\
DCT-Seg \cite{peng2020deep} & 78.20 (0.70) & 83.11 (0.20) & 90.22 (0.24) & 83.84 (0.10)\\
\re{ADVNET \cite{vu2019advent}} & \re{73.85 (1.29)} & \re{74.92 (0.85)} & \re{86.12 (0.53)} & \re{78.30 (0.87)}\\
\re{MT \cite{tarvainen2017mean}} & \re{82.99 (0.49)} & \re{80.43 (1.02)} & \re{89.33 (0.33)} & \re{84.25 (0.56)}\\
\re{Mut. Info. \cite{peng2020mutual}} & \re{81.98 (0.62)} & \re{75.75 (0.47)} & \re{87.89 (0.11)} & \re{81.87 (0.32)}\\
\re{Cons. Reg. \cite{peng2020mutual}} & \re{82.30 (0.60)} & \re{79.43 (0.81)} & \re{88.55 (0.37)} & \re{83.42 (0.48)}\\
\re{MSE \cite{peng2020mutual}} & \re{82.82 (0.35)} & \re{79.91 (0.72)} & \re{88.84 (0.77)} & \re{83.85 (0.20)}\\
\re{KL \cite{peng2020mutual}} & \re{\textbf{85.08 (0.10)}} & \re{81.08 (0.42)} & \re{{90.72 (0.44)}} & \re{85.63 (0.20)}\\
\re{Peng \etal \cite{2021Boosting}} & \re{81.87 (0.54)} & \re{83.65 (0.26)} & \re{\textbf{91.76 (0.32)}} & \re{85.76 (0.16)}\\
\textbf{Ours} & 82.69 (0.48) & \textbf{84.87 (0.55)} &  {90.74 (0.84)} & \textbf{86.10 (0.62)}\\
\midrule
\textit{Fully supervised} & 88.98 (0.09) & 84.95 (0.15) & 92.44 (0.33) & 88.79 (0.13) \\
\bottomrule
\end{tabular}
\end{table}

\begin{figure}[htbp]
\centering
\subfigure{
\begin{minipage}[t]{0.2\linewidth}
\centering
\includegraphics[width=0.7in]{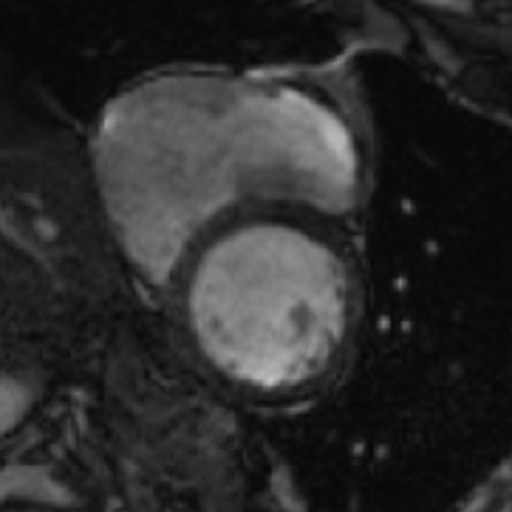}
\end{minipage}}
\subfigure{
\begin{minipage}[t]{0.2\linewidth}
\centering
\includegraphics[width=0.7in]{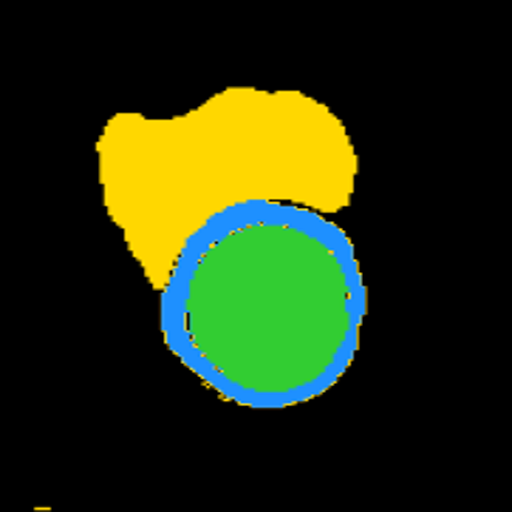}
\end{minipage}}
\subfigure{
\begin{minipage}[t]{0.2\linewidth}
\centering
\includegraphics[width=0.7in]{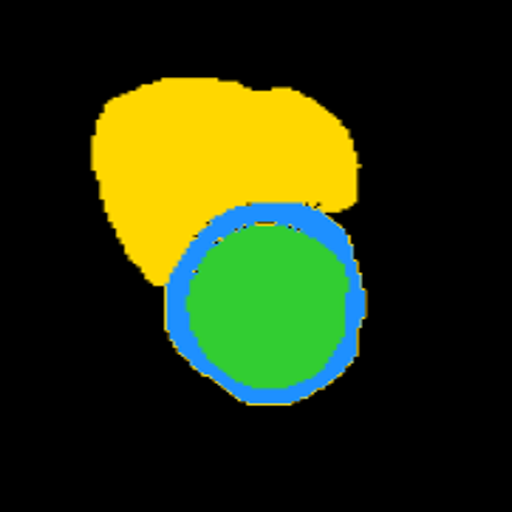}
\end{minipage}}
\subfigure{
\begin{minipage}[t]{0.2\linewidth}
\centering
\includegraphics[width=0.7in]{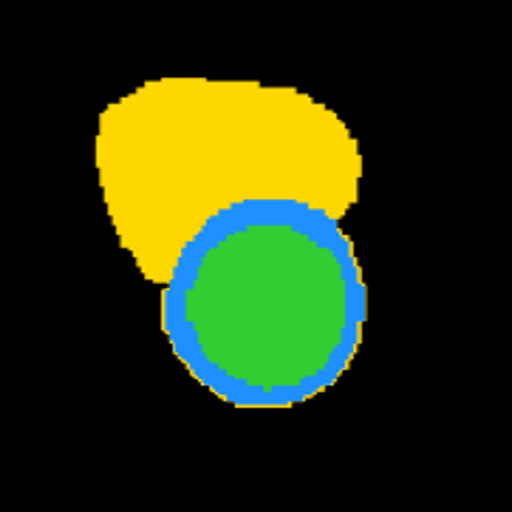}
\end{minipage}}
\\
\vspace*{-0.3cm}
\subfigure{
\begin{minipage}[t]{0.2\linewidth}
\centering
\includegraphics[width=0.7in]{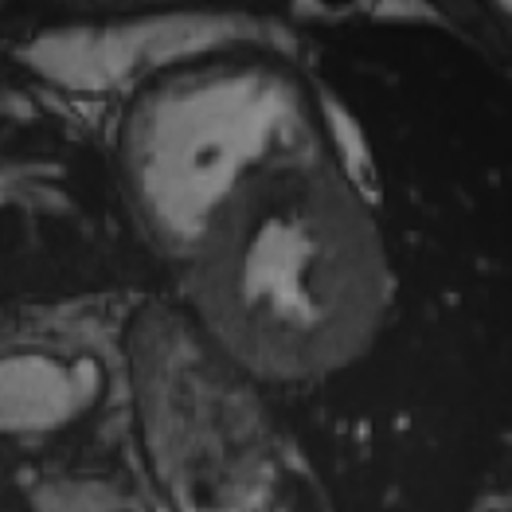}
\end{minipage}}
\subfigure{
\begin{minipage}[t]{0.2\linewidth}
\centering
\includegraphics[width=0.7in]{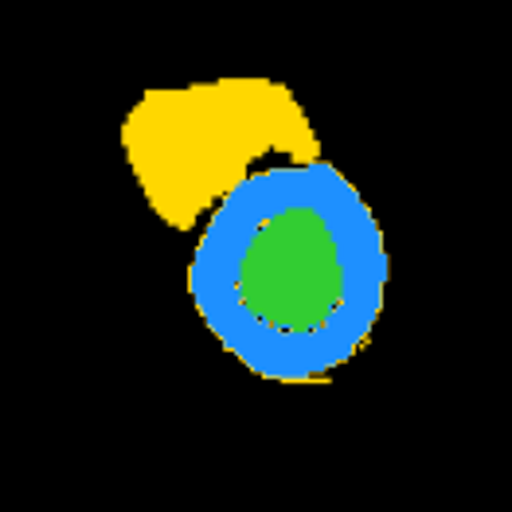}
\end{minipage}}
\subfigure{
\begin{minipage}[t]{0.2\linewidth}
\centering
\includegraphics[width=0.7in]{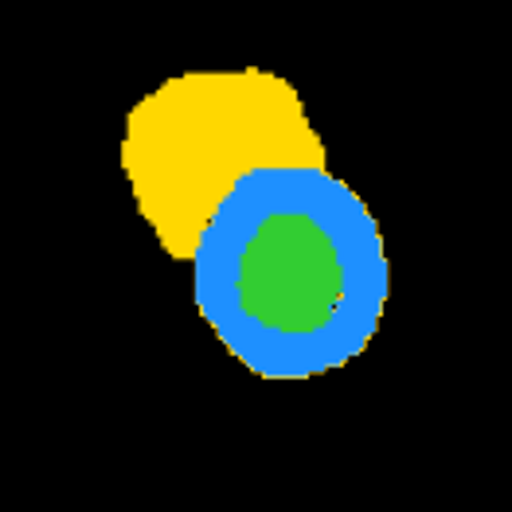}
\end{minipage}}
\subfigure{
\begin{minipage}[t]{0.2\linewidth}
\centering
\includegraphics[width=0.7in]{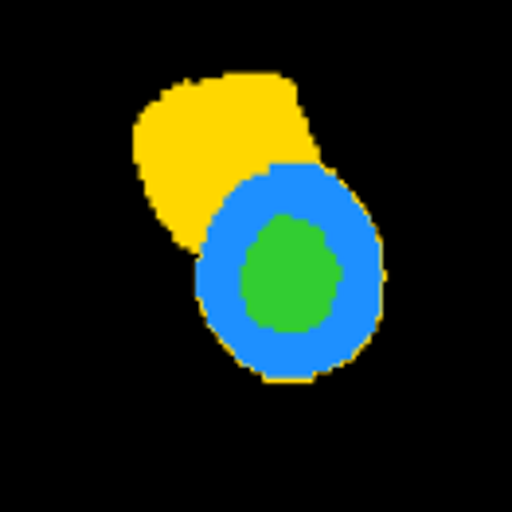}
\end{minipage}}
\\
\vspace*{-0.3cm}
\subfigure{
\begin{minipage}[t]{0.2\linewidth}
\centering
\includegraphics[width=0.7in]{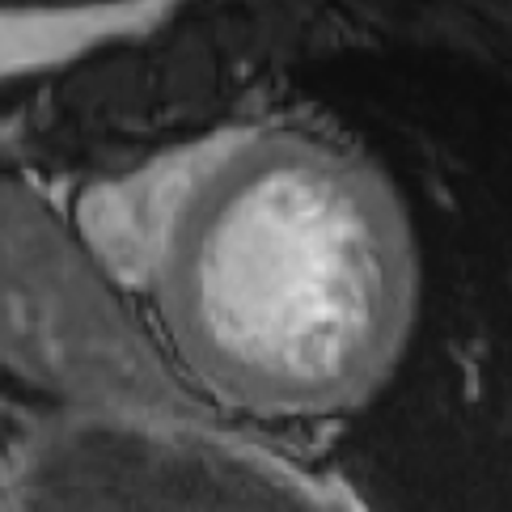}
\end{minipage}}
\subfigure{
\begin{minipage}[t]{0.2\linewidth}
\centering
\includegraphics[width=0.7in]{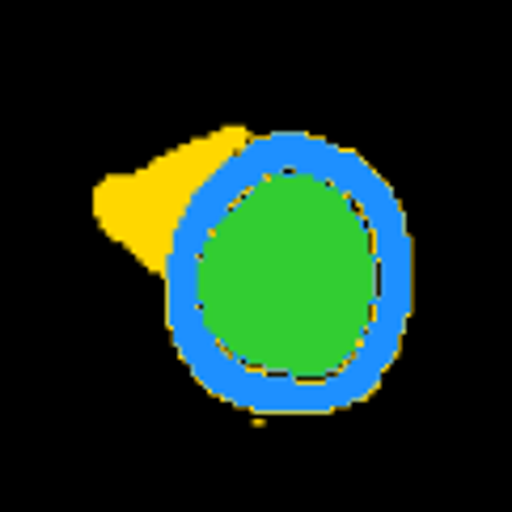}
\end{minipage}}
\subfigure{
\begin{minipage}[t]{0.2\linewidth}
\centering
\includegraphics[width=0.7in]{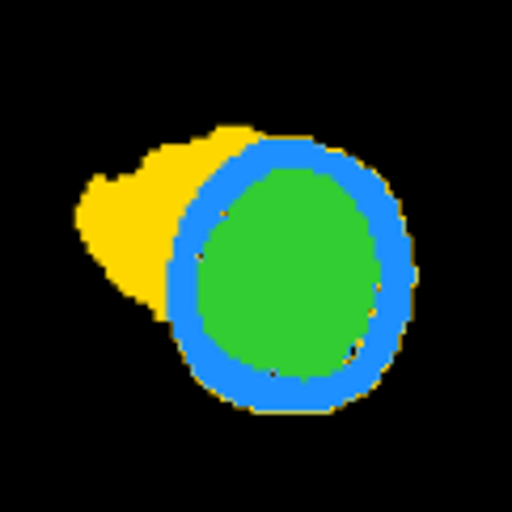}
\end{minipage}}
\subfigure{
\begin{minipage}[t]{0.2\linewidth}
\centering
\includegraphics[width=0.7in]{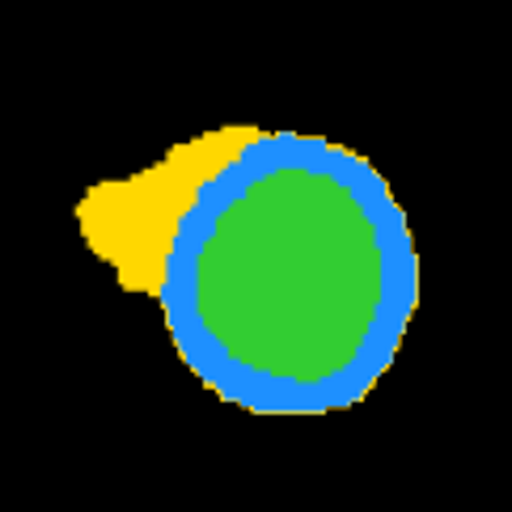}
\end{minipage}}
\\
\vspace*{-0.3cm}
\subfigure{
\begin{minipage}[t]{0.2\linewidth}
\centering
\includegraphics[width=0.7in]{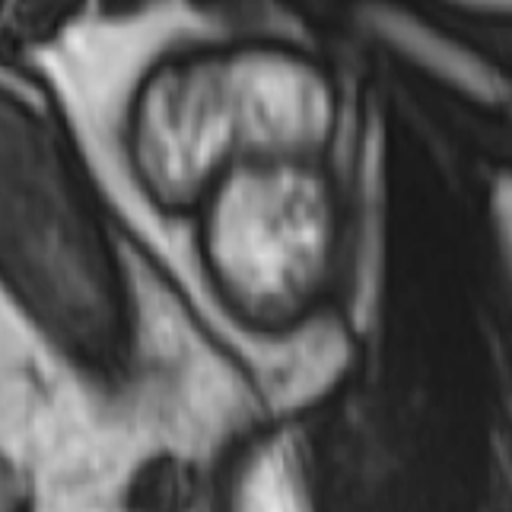}
\centerline{\scriptsize{Original image}}
\end{minipage}}
\subfigure{
\begin{minipage}[t]{0.2\linewidth}
\centering
\includegraphics[width=0.7in]{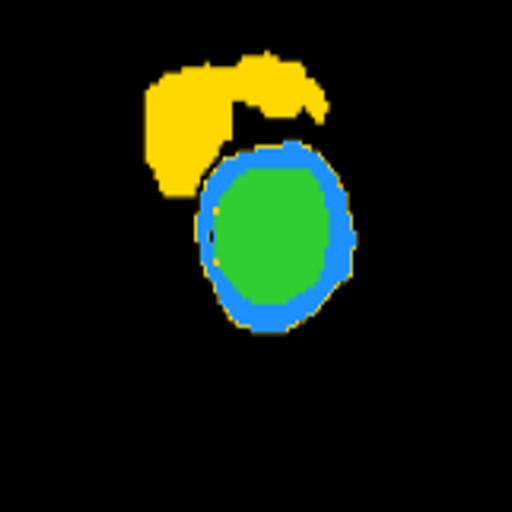}
\centerline{\scriptsize{Mean teacher}}
\end{minipage}}
\subfigure{
\begin{minipage}[t]{0.2\linewidth}
\centering
\includegraphics[width=0.7in]{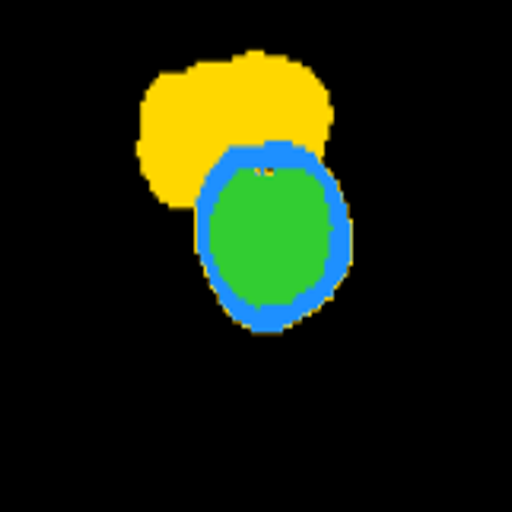}
\centerline{\scriptsize{Ours}}
\end{minipage}}
\subfigure{
\begin{minipage}[t]{0.2\linewidth}
\centering
\includegraphics[width=0.7in]{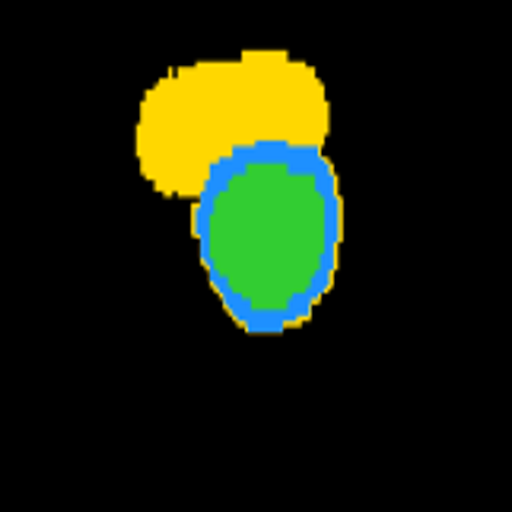}
\centerline{\scriptsize{Ground truth}}
\end{minipage}}
\caption{The visual segmentation examples of mean teacher \cite{tarvainen2017mean} and our method on ACDC segmentation.}\label{fig:visual_acdcexample}
\end{figure}

We also present several visualized segmentation examples of mean teacher and our method in Figure \ref{fig:visual_acdcexample} for comparison. Different colors indicate different classes for segmentation. Compared with mean teacher, our method also obtains more precise segmentation on some difficult boundary regions. 




\section{Conclusion}
\label{sec:conclusion}
How to estimate and handle uncertainty remains a crucial problem in semi-supervised medical image segmentation.
In this paper, we proposed a novel method of estimating uncertainty by capturing the inconsistent prediction between multiple cost-sensitive settings. Our definition of uncertainty directly relies on the classification output without requiring any predefined boundary-aware assumption.
Based on this definition, we also presented a separate self-training strategy to treat certain and uncertain regions differently. To achieve end-to-end training, three components, \ie, CRM, C-SN and UC-SN were developed to train our semi-supervised segmentation model.  
By extensively evaluating the proposed method in various medical image segmentation tasks, \eg, CT pancreas, MR endocardium, and multi-class ACDC segmentation, we demonstrated that our method outperforms other related baselines in terms of segmentation accuracy.

\section*{Acknowledgement}
The authors would like to thank Dr. Wenjuan Xie (NNU), Lihe Yang (NJU), and Zhen Zhao (USYD $\&$ NJU) for discussion and comments. The work of Yinghuan Shi, Jian Zhang, and Tong Ling was supported by National Key Research and Development Program of China (2019YFC0118300), and Lei Qi  was  supported  by  China  Postdoctoral  Science  Foundation Project (2021M690609) and Jiangsu Natural Science Foundation Project (BK20210224).

{\small
\bibliographystyle{ieeetrans}
\bibliography{egbib}
}
\end{document}